\documentclass[letterpaper]{article} 
\usepackage{aaai25}  
\usepackage{times}  
\usepackage{helvet}  
\usepackage{courier}  
\usepackage[hyphens]{url}  
\usepackage{graphicx} 
\usepackage{natbib}  
\usepackage{caption} 
\usepackage{amsmath}
\usepackage{algorithm}
\usepackage{algorithmic}
\usepackage{rotating}
\usepackage{graphics, bm, physics}
\usepackage{amssymb}
\usepackage{amsthm}
\usepackage{thmtools}
\usepackage{thm-restate}
\usepackage{cleveref}

\usepackage{nicefrac}
\usepackage{subfigure}
\newtheorem{definition}{Definition}
\newtheorem{assumption}{Assumption}

\usepackage[table]{xcolor}
\usepackage{colortbl}
\usepackage{tcolorbox}
\usepackage{tikz}

\usepackage{newfloat}
\usepackage{listings}
\DeclareCaptionStyle{ruled}{labelfont=normalfont,labelsep=colon,strut=off} 
\floatstyle{ruled}
\newfloat{listing}{tb}{lst}{}
\floatname{listing}{Listing}
\title{On Effects of Steering Latent Representation for \\Large Language Model Unlearning}
\author{Dang Huu-Tien\textsuperscript{\rm 1}, Tin Pham\textsuperscript{\rm 1}, Hoang Thanh-Tung\textsuperscript{\rm 2}, and Naoya Inoue\textsuperscript{\rm 1,\rm 3}}

\affiliations{
\textsuperscript{\rm 1}Japan Advanced Institute of Science and Technology\\
    \textsuperscript{\rm 2}VNU University of Engineering and Technology, Vietnam\\
    \textsuperscript{\rm 3}RIKEN\\
    \textsuperscript{\rm 1}\{s2310417, tinpham, naoya-i\}@jaist.ac.jp, \textsuperscript{\rm 2}htt210@gmail.com
}
\usepackage{bibentry}

\begin{document}

\maketitle

\begin{abstract}
Representation Misdirection for Unlearning (RMU), which steers model representation in the intermediate layer to a target random representation, is an effective method for large language model (LLM) unlearning. Despite its high performance, the underlying cause and explanation remain underexplored.
In this paper, we theoretically demonstrate that steering forget representations in the intermediate layer reduces token confidence, causing LLMs to generate wrong or nonsense responses. We investigate how the coefficient influences the alignment of forget-sample representations with the random direction and hint at the optimal coefficient values for effective unlearning across different network layers. We show that RMU unlearned models are robust against adversarial jailbreak attacks.
Furthermore, our empirical analysis shows that RMU is less effective when applied to the middle and later layers in LLMs.
To resolve this drawback, we propose \textit{Adaptive RMU}---a simple yet effective alternative method that makes unlearning effective with most layers.
Extensive experiments demonstrate that Adaptive RMU significantly improves the unlearning performance compared to prior art while incurring no additional computational cost.
\end{abstract}

%

\section{Introduction}
LLMs achieved remarkable performance through pre-training on large amounts of internet texts and rigorous alignment processes for safety enhancement. 
Despite the immense effort in safety research, LLMs are still vulnerable to adversarial jailbreak attacks and can exhibit unwanted behaviors~\cite{shahscalable, zou2023universal, jones2023automatically, yuan2024gpt, wei2024jailbroken}.

Machine Unlearning~\cite{7163042, bourtoule2021machine, nguyen2022survey, 10.1145/3603620, liu2024machine} has emerged as a promising method for mitigating unforeseen risks in LLMs before deployment.
\citet{wmdp} introduced Representation Misdirection for Unlearning (RMU)---an unlearning method that steers the representations of forget-samples (\textit{i.e.} samples that the model should forget) toward a random representation while keeping the representations of retain-samples (\textit{i.e.} samples that the model should remember) unchanged.
RMU significantly degrades models' accuracy on forget-tasks, while only slightly affecting the performance on retain-tasks and demonstrates stronger robustness against adversarial jailbreak attacks. 
However, the reason for RMU's effectiveness is not well understood, hindering the development of better unlearning algorithms. In this paper, we make the following contributions:
\begin{itemize}
    \item We theoretically analyze the impact of the RMU method on LLM unlearning.
    \item We investigate the connection between RMU and adversarial robustness. We demonstrate that RMU impedes the adversary's ability to determine optimal updates for generating adversarial samples, thus improving the adversarial robustness of the unlearned model.
    
    \item We empirically show that the RMU forget loss, which minimizes the mean squared error (MSE) between forget representation and a fixed scaled random vector, fails to converge when the norm of the forget representation is larger than the scaling coefficient, making RMU less effective when applied to middle and last layers in LLMs.
    \item To overcome RMU's limitation, we introduce \emph{Adaptive RMU}---a variant that adaptively adjusts the coefficient value based on the norm of the forget representation. Experimental results show that Adaptive RMU achieves higher drop-in-accuracy for forget knowledge, maintaining high performance on general knowledge, and enables effective unlearning for most layers without incurring additional computational overhead.
\end{itemize}
\section{Background and Related Work}

\paragraph{Machine Unlearning.}  
A natural unlearning approach is leave-some-out retraining: retraining the model from scratch without the forget samples. However, this method becomes more computationally expensive as the size of datasets and modern deep networks grows. Existing works focus on approximating unlearning~\cite{warnecke2021machine, izzo2021approximate, sekhari2021remember, isonuma2024unlearning} using influence function~\cite{koh2017understanding, grosse2023studying}, gradient ascent~\cite{thudi2022unrolling}, second-order approximation~\cite{jia2024soul}, negative preference optimization~\cite{zhang2024negative}, and embedding corrupted~\cite{liu2024large}. Other views on the landscape of machine unlearning include: unlearning in text classification~\cite{ma2022learn}, image classification and recognition~\cite{ginart2019making, golatkar2020eternal, fan2024salun, choi2023towards, cha2024learning}, image-to-image generative models~\cite{li2024machine}, diffusion models~\cite{gandikota2023erasing, zhang2024forget, kumari2023ablating,bui2024adversarial}, multimodal unlearning~\cite{cheng2023multimodal}, federated unlearning~\cite{romandini2024federated, 10.1145/3485447.3512222, 10.5555/3618408.3618577, halimi2022federated, jeong2024sok}, graph unlearning~\cite{chen2022graph, chien2023efficient, 10.1145/3580305.3599271, cheng2023gnndelete, dukler2023safe, zhu2023heterogeneous, li2024towards, tan2024unlink}, recommender systems~\cite{zhang2023recommendation, chen2024post, li2023making, wang2025towards}, certified minimax unlearning~\cite{liu2024certified}, targeted types of unlearning information~\cite{cooper2024machine}, and evaluation on unlearning~\cite{lynch2024eight, hayes2024inexact, shi2024detecting, shi2024muse}.
\paragraph{LLM Unlearning.} Due to the large size of the parameters and training data, LLM poses a new challenge to unlearning. Recent studies in LLM unlearning mainly focus on task or context-specific settings such as unlearning copyrighted material from the Harry Potter series~\cite{eldan2023s}, in-context unlearning~\cite{pawelczykcontext}, fictitious unlearning~\cite{maini2024tofu}, specific harmful input-output~\cite{llmu, liu2024towards}, sensitive and private information~\cite{jang-etal-2023-knowledge, wu-etal-2023-depn, patil2024can}, gender bias~\cite{belrose2023leace} or concepts~\cite{hong2024intrinsic, bui2024adversarial}. More recently, \citet{wmdp} consider unlearning an entire distribution of hazardous knowledge given limited samples.
\paragraph{Notation \& problem formulation.} 
Let $\mathcal{D}_{\text{forget}}$ and $\mathcal{D}_{\text{retain}}$ be the forget and retain sets, respectively. 
Let $f_{\theta}: \mathbb{R}^{n\times d} \mapsto \mathbb{R}^{n\times |V|}$ be an autoregressive LLM parameterized by $\theta$ that maps a prompt input $x_{1:n}$ consisting of $n$ tokens $\{x_1,x_2,...,x_n\}$ to an output of probability distributions over the vocabulary $V$. We denote $h_{\theta}^{(l)}(x)$ the \textit{averaged} hidden states of all tokens in $x_{1:n}$ obtained from the $l$-th layer of $f_{\theta}$. For simplicity, throughout this paper, we use $h^{(l)}(x)$ to present $h_{\theta}^{(l)}(x)$. For operators, we denote $\circ$ as the decomposition operator, and $||\cdot||$ is the Euclidean norm. Our goal is to unlearn the undesired harmful knowledge $\mathcal{D}_{\textnormal{forget}}$ from $f_{\theta}$ while retaining general knowledge 
$\mathcal{D}_{\textnormal{retain}}$. 
Unlearned models should be robust to knowledge recovery attacks that attempt to recover harmful knowledge from the model. 
\paragraph{Representation Misdirection for Unlearning} (RMU;~\citet{wmdp})  is a fine-tuning based unlearning method inspired by representation engineering~\cite{zou2023representation} that steers the model's representation of forget samples $x_F \in \mathcal{D}_\text{forget}$ to a random vector and regularizes the model representation of retain samples $x_R \in \mathcal{D}_\text{retain}$ back to the original model representation, by optimizing the MSE loss:
\begin{align}
    \mathcal{L} &= \mathbb{E}_{x_F\in\mathcal{D}_{\text{forget}}}||h_{\theta^{\text{unlearn}}}^{(l)}(x_F)-c\bm u||_2^2 \nonumber\\&+ \alpha \mathbb{E}_{x_R\in\mathcal{D}_{\text{retain}}}||h_{\theta^{\text{unlearn}}}^{(l)}(x_R)-h_{\theta^{\text{frozen}}}^{(l)}(x_R)||_2^2,\label{eq1}
\end{align}
where $\theta^{\text{unlearn}}$ and $\theta^{\text{frozen}}$ are parameters of the update model and frozen model respectively, $\bm u$ is a fixed random unit vector where each element is sampled from Uniform distribution $U(0,1)$, $c \in \mathbb{R}$ is a fixed scaling coefficient and $\alpha \in \mathbb{R}$ is a retain weight. RMU updates $\theta^{\text{unlearn}}$ toward the direction of the gradient of the loss $\mathcal{L}$ using gradient descent.

\section{Theoretical Analysis}

\subsection{The Confidence of Tokens Generated by RMU Models}
\label{sec:3.1}
In general, samples from the shifted distribution (such as wrong label or out-of-distribution) are associated with smaller ``confidence'' scores such as softmax probability \cite{hendrycks2017a, northcutt2021confident}, maximum logit~\cite{pmlr-v162-hendrycks22a, wei2022mitigating}, $\ell^2$-distance~\cite{sun2022out}, energy score~\cite{liu2020energy}, and cosine similarity~\cite{ngoc2023cosine}. Recently, LLM has shown a tendency to produce a lower (higher) confidence in its incorrect (correct) answers in multiple-choice Q\&A~\cite{plaut2024softmax}. Building on previous works, we hypothesized that the \textit{logit} of generated tokens by RMU models exhibit randomness. As seen by a deep network, such randomization signifies low confidence in the logit, resulting in nonsensical or incorrect responses. To validate the hypothesis, we conducted an analysis of the logits of generated tokens produced by RMU models. To facilitate subsequent analysis, we make the following definition and assumption.

\begin{definition}
\label{def1}
(Unlearned model \& logit of forget-tokens on unlearned model). Let $f^{(l:k)} = g^{(l:k)} \circ h^{(l)}$, where $g^{(l:k)}$ be the transformation from layer $l$ to layer $k$ of network $f$, for any two layers $k > l$; $l\in [1...L]$. We define the unlearned model $f^{\textnormal{unlearn}} = \bm W( f^{(l:L),
\textnormal{steered}})=\bm W(g^{(l:L)}\circ h^{(l),\textnormal{steered}})$, $h^{(l), \textnormal{steered}}$ is the steered representation of the given input at layer $l$ and $\bm W$ is the unembedding matrix which maps output hidden states back to the vocabulary space. Given a forget input $x_{F,1:n}$, the logit of the next token $x_{F,n+1}$ obtained from unlearned model $f^{\text{unlearn}}$ is defined as:
\begin{align}
    f^{\text{unlearn}}(&x_{F,n+1}|x_{F,1:n}) = \bm W f^{(l:L),\text{steered}}(x_{F,n+1}|x_{F,1:n})\nonumber\\
    &=\bm W(g^{(l:L)}\circ h^{(l),\textnormal{steered}})(x_{F,n+1}|x_{F,1:n})\nonumber\\
    &=\bm Wg^{(l:L)}(h^{(l),\textnormal{steered}}(x_{F,n+1}|x_{F,1:n}))\label{eq2}
\end{align}
\end{definition}
\begin{assumption} A well-unlearned model shifts the representation of all tokens in a forget-sample $x_{F,1:n}$ at layer $l$ to a scaled random vector $c\bm u$. More concretely,
\begin{align}
    h^{(l),\textnormal{steered}}(x_{F,i}) = c \bm u + \bm\epsilon, \label{eq3}
\end{align}
where $x_{F,i}$ is the $i$-th token in $x_F$, $\bm{\epsilon}$ is a small error. Without losing generality, we assume that $\bm \epsilon$ is sampled from Normal distribution $\mathcal{N}(\bm 0, \eta \bm I)$, where $\eta \bm I$ is the covariance matrix, $\eta \in \mathbb{R}$.
\label{assumption1}
\end{assumption} 
\begin{restatable}{proposition}{gold}
\label{theorem1}
If Assumption \ref{assumption1} holds, by Definition \ref{def1}, the logit value of forget token $x_{F,n+1}$ generated by unlearned model $f^{\textnormal{unlearn}}$ given as $f^{\textnormal{unlearn}}(x_{F,n+1}|x_{F,1:n})$
follows the Normal distribution 
    $\mathcal{N}\left(\bm Wg^{(l:L)}(\bm z), \eta\bm W\nabla_{\bm z}g^{(l:L)}(\bm z)^{\top}\nabla_{\bm z}g^{(l:L)}(\bm z)\bm W^{\top}\right),\label{eq5}$
where $\bm z = c \bm u$.
\end{restatable}

\begin{proof} Assumption~\ref{assumption1} implies that in a well-unlearned model, token $x_{F,n+1}$ is independent of the previous tokens, thus we have:
\begin{align}h^{(l),\textnormal{steered}}(x_{n+1}|x_{F, 1:n}) \approx h^{(l),\text{steered}}(x_{F,n+1})= c\bm u + \bm \epsilon \label{eq4}
\end{align} 
Denote $\bm z = c\bm u$. Substituting Eqn.~\ref{eq4} into Eqn.~\ref{eq2}, we get:
\begin{align}
{f^{\text{unlearn}}}(x_{F,n+1}|x_{F,1:n}) \approx \bm Wg^{(l:L)}(\bm z + \bm\epsilon)
\end{align}
Since $\bm \epsilon$ is small, we approximate the function $g^{(l:L)}(\bm z + \bm\epsilon)$ by its first-order derivative:
\begin{align}f^{\text{unlearn}}(x_{F,n+1}|x_{F 1:n})\approx \bm W(g^{(l:L)}(\bm z) + \nabla_{\bm z}g^{(l:L)}(\bm z)^{\top}\bm \epsilon)\end{align}
Given that $\bm \epsilon \sim \mathcal{N}(\bm 0, \eta\bm I)$, by applying the affine transformation property of the multivariate normal distribution, we get:
\begin{align}
&f^{\textnormal{unlearn}}(x_{F,n+1}|x_{F,1:n}) \nonumber\\&\sim \mathcal{N}\left(\bm Wg^{(l:L)}(\bm z), \eta\bm W\nabla_{z}g^{(l:L)}(\bm z)^{\top}\nabla_{z}g^{(l:L)}(\bm z) \bm W^{\top}\right)
\end{align}
Since $\bm u \sim U(0,1)$, then $\bm z \sim U(0,c)$.
By definition of variance, we have: $\textnormal{Var}(\bm z) =\textnormal{Var}(c\bm u) = c^2\textnormal{Var}(\bm u)$.
\end{proof}
Proposition \ref{theorem1} suggests that the variance of $f^{\text{unlearn}}(x_{F,n+1}|x_{F,1:n})$ is controlled by (i) $\eta$: a scalar variance and (ii) $\bm W\nabla_{\bm z}g^{(l:L)}(\bm z)^{\top}\nabla_{\bm z}g^{(l:L)}(\bm z)\bm W^{\top}$: the product of $\bm W\nabla_{\bm z}g^{(L)}(\bm z)^{\top}$ and $\nabla_{\bm z}g^{(L)}(\bm z)\bm W^{\top}$.
If $f^{\text{unlearn}}(x_{F,n+1}|x_{F,1:n})$ has high variance, the logit values are more random. Since $\bm\epsilon$ presents a small error, then $\bm\epsilon$ varies for different inputs $x_F$. This variation makes it difficult to control the variance of the logit by $\eta$. The main effect depend on $\bm W\nabla_{\bm z}g^{(l:L)}(\bm z)^{\top}\nabla_{\bm z}g^{(l:L)}(\bm z)\bm W^{\top}$. While the unembedding matrix $\bm W$ is unchanged after unlearning, the product $\nabla_{\bm z}g^{(l:L)}(\bm z)^{\top}\nabla_{\bm z}g^{(l:L)}(\bm z)$ varies depending on the specific characteristics of sub-networks $g^{(l:L)}$ and input $\bm z = c\bm u$. Unfortunately, $g^{(l:L)}$ is a composition of transformer layers, which is highly nonlinear, making it difficult to have a complete analysis. The variance of $\bm z$, derived as $\text{Var}(\bm z) = c^2\text{Var}(\bm u)$, is proportional to $c$; \textit{i.e.} when $c$ gets larger, the variance of $\bm z$ is higher. This could increase the variability of $g^{(l:L)}(\bm z)$ and the gradient $\nabla_{\bm z}g^{(l:L)}(\bm z)$. \textit{A larger $c$ could introduces more randomness to the logit}.  We conduct an empirical analysis to understand the confidence of generated tokens by RMU models in Section 4.1.

\subsection{The Effect of Coefficient $c$ on Forget-sample Representations}
\label{sec:3.2}
RMU forget loss steers forget-sample representation $h^{(l)}(x_F)$ aligns with a random direction given by $\bm u$ and scales the magnitude of $h^{(l)}(x_F)$ to $c$ (Eqn~\ref{eq1}). While vector $\bm u$ is predetermined before unlearning, the magnitude of $h^{(l)}(x_F)$ varies depending on input $x_F$ and specific properties of layer $l$. This raises the following research questions:

\noindent RQ1 (Direction): \textit{``How does the coefficient $c$ influence the alignment between $h^{(l)}(x_F)$ with $\bm u$.''}\\
RQ2 (Magnitude): \textit{``What is the optimal value of the coefficient $c$ for effectively unlearning with different layers.''}

\paragraph{Unlearning as minimizing the noise sensitivity.} We aim to answer these questions by analyzing the unlearning problem under a noise compression view. We consider the output of a transformation $f^{(l:k)}$ on input $x$: $f^{(l:k)}(x) = (g^{(l:k)} \circ h^{(l)})(x) = g^{(l:k)} \left(h^{(l)}(x)\right)$. Suppose we compress a noise vector $\bm \xi$ to the representation $h^{(l)}$ of layer $l$ at input $x$, then the output become $g^{(l:k)} \left(h^{(l)}(x) + \bm \xi\right)$. Naturally, if layer $g^{(l:k)}$ is robust (less sensitive) to noise $\bm \xi$, then $\bm \xi$ has a small effect on the output of $g^{(l:k)}$ \textit{i.e.} the normalized squared norm 
\begin{align}
        \Phi(g^{(l:k)}, x) = \frac{||g^{(l:k)}\left(h^{(l)}(x)+ \bm\xi\right) - g^{(l:k)} \left(h^{(l)}(x)\right)||^2}{||g^{(l:k)} \left(h^{(l)}(x)\right)||^2}
    \end{align}
is small. In contrast, a higher $\Phi(g^{(l:k)},x)$ mean $g^{(l:k)}$ is higher sensitive to noise $\bm \xi$ at input $x$. For a dataset $\mathcal{D}_{\text{forget}}$, we define the \textit{noise sensitivity} of a layer $g^{(l:k)}$ w.r.t $\bm\xi$ on $\mathcal{D}_{\text{forget}}$ as:
\begin{align}
        &\Phi(g^{(l:k)}, \mathcal{D}_{\text{forget}}) \nonumber\\&=\frac{||g^{(l:k)}(\hat{h}^{(l)}(x_F)+ \bm\xi) - g^{(l:k)} (\hat h^{(l)}(x_F))||^2}{||g^{(l:k)} (\hat h^{(l)}(x_F))||^2},
    \end{align}
    where $\hat h^{(l)}(x_F)$ is the mean of $h^{(l)}(x_F)$ over $x_F\in \mathcal{D}_{\text{forget}}$.
During unlearning, RMU steers $h^{(l)}(x_F)$ for all $x_F\in \mathcal{D}_{\text{forget}}$ to the fixed vector $c\bm u + \bm\epsilon$ \textit{i.e.} $||g^{(l:k)}(c\bm u + \bm \epsilon) - g^{(l:k)}(\hat h^{(l)}(x_F))||^2$ is minimized. If we let $\bm \xi = c \bm u + \bm \epsilon- \hat h^{(l)}(x_F)$, we can define the unlearning problem as minimizing the noise sensitivity of the layer. This objective is described by
\begin{align}
\min \frac{||g^{(l:k)}(c\bm u +\bm\epsilon) - g^{(l:k)} (\hat h^{(l)}(x_F))||^2}{||g^{(l:k)} (\hat h^{(l)}(x_F))||^2}
\end{align}
While $g^{(l:k)}$ is a composition of transformer layers, which is hard to expand it in term of $c$. Therefore, we propose to use the Jacobian matrix $\bm J^{(l:k)}(x_F)$---a linearized of $g^{(l:k)}$ at $x_F$---which describes the change in the output of $g^{(l:k)}$ due to a noise perturbed in the input $\hat h^{(l)}(x_F)$. For simplification, we write $\hat h^{(l)}$, $\bm J^{(l:k)}$ instead of $\hat h^{(l)}(x_F)$, $\bm J^{(l:k)}(x_F)$ respectively. The objective becomes
\begin{align}
\min\frac{||\bm J^{(l:k)}(c\bm u +\bm \epsilon) - \bm J^{(l:k)} \hat h^{(l)}||^2}{||\bm J^{(l:k)} \hat h^{(l)}||^2} \label{eq11}
\end{align}
Since $\bm J^{(l:k)}$ is a linear transformation, then 
\begin{align}
    ||\bm J^{(l:k)}(c\bm u + \bm\epsilon) - \bm J^{(l:k)} \hat h^{(l)}||^2 = ||\bm J^{(l:k)}(c\bm u +\bm\epsilon - \hat h^{(l)})||^2
\end{align}
Let $\bm v = \bm\epsilon - \hat h^{(l)}$. By definition of the squared norm, we have:
\begin{align}
    ||\bm J^{(l:k)}(c\bm u + \bm v)||^2 &= (\bm J^{(l:k)}(c\bm u +\bm v))^{\top} \bm J^{(l:k)}(c\bm u +\bm v)
    \nonumber\\&=  (c\bm u + \bm v)^{\top}\bm{J}^{(l:k)\top}\bm J^{(l:k)}(c\bm u +\bm v)\label{eq13}
\end{align}
Let matrix $\bm A = \bm{J}^{(l:k)\top} \bm J^{(l:k)}$. Expand the right-hand side of Eqn.~\ref{eq13}, we get:
\begin{align}
     &||\bm J^{(l:k)}(c\bm u + \bm v)||^2 \nonumber\\&= (c\bm u)^{\top}\bm A c\bm u + (c\bm u)^{\top}\bm A \bm v + \bm v^{\top}\bm A c\bm u + \bm v^{\top}\bm A \bm v \label{eq14}
\end{align}
Since $\bm A$ is a symmetric matrix (\textit{i.e.} $\bm A^{\top} = \bm A$), then
\begin{align}
    (c\bm u)^{\top}\bm A \bm v = (c\bm u)^{\top} \bm A^{\top} \bm v = (\bm A c\bm u)^{\top} \bm v = \bm v^{\top} \bm A c\bm u \label{eq15}
\end{align}
Substituting $ (c\bm u)^{\top}\bm A \bm v = \bm v^{\top} \bm A c\bm u$ into Eqn.~\ref{eq14} we get:
\begin{align}
    ||\bm J^{(l:k)}(c\bm u + \bm v)||^2 = c^2\bm u^{\top}\bm A \bm u + 2c \bm u^{\top}\bm A \bm v + \bm v^{\top}\bm A \bm v \label{eq16}
\end{align}
Substituting Eqn.~\ref{eq16} into Eqn.~\ref{eq11}, the objective becomes
\begin{align}
 \min \frac{c^2\bm u^{\top}\bm A \bm u + 2c \bm u^{\top}\bm A \bm v + \bm v^{\top}\bm A \bm v}{||\bm J^{(l:k)}\hat{h}^{(l)}||^2}\label{eq17}
\end{align}
Taking its derivative w.r.t $c$ and set it to zero:
\begin{align}
    \frac{2 \bm u^{\top}\bm A \bm u c + 2 \bm u^{\top}\bm A \bm v}{||\bm J^{(l:k)}\hat{h}^{(l)}||^2} = 0
\end{align}
Since $||\bm J^{(l:k)}\hat{h}^{(l)}||^2$ is not zero, solve for $c$:
\begin{align}
    c &= -\frac{\bm u^{\top}\bm A \bm v}{\bm u^{\top} \bm A \bm u} = \frac{\bm u^{\top}\bm{J}^{(l:k)\top}\bm J^{(l:k)} (\hat h^{(l)} - \bm \epsilon)}{\bm u^{\top}\bm{J}^{(l:k)\top}\bm J^{(l:k)}\bm u} \nonumber\\&= \frac{(\bm J^{(l:k)} \bm u)^{\top}\bm J^{(l:k)}(\hat h^{(l)} - \bm \epsilon)}{||\bm J^{(l:k)} \bm u||^2}\nonumber\\&= \frac{||\bm J^{(l:k)} (\hat h^{(l) }- \bm \epsilon))||}{||\bm J^{(l:k)} \bm u||}\cos(\bm J^{(l:k)} \bm u, \bm J^{(l:k)} (\hat h^{(l)}-\bm\epsilon))
\end{align}
Since $\frac{||\bm J^{(l:k)} (\hat h^{(l)}-\bm\epsilon)||}{||\bm J^{(l:k)} \bm u||}$ is positive, then \textit{$c$ and $\cos(\bm J^{(l:k)} \bm u, \bm J^{(l:k)} (\hat h^{(l)}-\bm\epsilon))$ are positively correlated.}

This means smaller (larger) $c$ indicates less (more) \textit{alignment} between $\bm J^{(l:k)} \bm u$ and $\bm J^{(l:k)} (\hat h^{(l)}-\bm\epsilon)$. Given that the Jacobian $\bm J^{(l:k)}$ describes how small changes in the input lead to changes in the output using linear approximation around a given point. If $\bm J^{(l:k)}$ does not vary drastically, it will not significantly alter the directions of $\bm u$ and $\hat h^{(l)}-\bm\epsilon$. In such cases, $\bm J^{(l:k)}$ will have a small effect on directional alignment, preserving the relative angles between $\bm u$ and $\hat h^{(l)}-\bm\epsilon$. Here, reasonably, \textit{$\bm u$ and $\hat h^{(l)}$ are becoming more aligned as $c$ increases} since error $\bm\epsilon \to \bm 0$ as unlearning becomes more accurate. 
\begin{figure}[!t] 
\centering 
\includegraphics[width=0.48\textwidth]{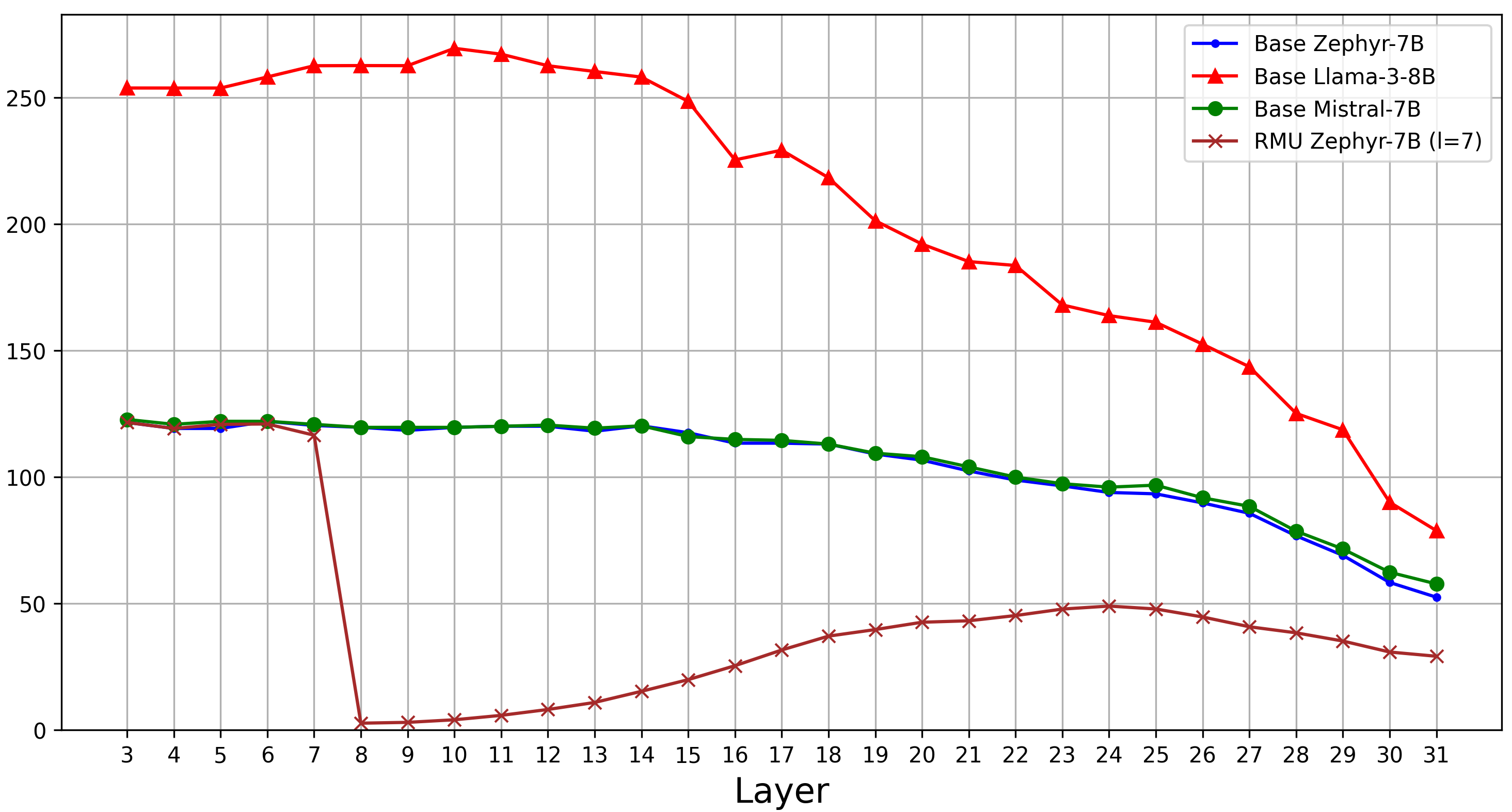}
\caption{Noise sensitivity of layer $g^{(l:k)}$, for $k \in [3...31]$ in base Zephyr-7B, base Llama-3-8B, base Mistral-7B, and RMU Zephyr-7B model. In the base models, a deeper layer has lower noise sensitivity, while the noise sensitivity is minimized in the RMU model (compress noise into $h^{(7)}$, the noise sensitivity of layer $k=8$ is minimized).}
\label{fig1}
\end{figure}

The above discussion does not directly address RQ2. However, the definition of the noise sensitivity suggests that the noise sensitivity of layer $g^{(l:k)}$ is characterized by the inherent properties of $g^{(l:k)}$, the representation $\hat h^{(l)}(x_F)$ (which is fixed) and the perturbed noise $\bm \xi$. If $\bm \xi$ is predetermined, the noise sensitivity of $g^{(l:k)}$ depends solely on its properties. This suggest the following experiment: we compute $\hat{h}^{(l)}(x_F)$---the mean of $h^{(l)}(x_F)$ over a set of input $x_F\in \mathcal{D}_{\text{forget}}$, compress a fix perturbed noise $\bm \xi$ into $\hat{h}^{(l)}(x_F)$. We then calculate the noise sensitivity of $g^{(l:k)}$ for different layers. Fig.~\ref{fig1} shows the noise sensitivity of layers across different models. We empirically observed that: \textit{the noise sensitivity decreases as layers go deeper and vary across different models}. Since noise sensitivity describes a layer's robustness to noise, higher noise sensitivity means $g^{(l:k)}$ requires smaller noise to produce the same level of output randomness, while lower noise sensitivity means it requires larger noise. In other words, early layers require smaller noise $\bm\xi$ (smaller $c$) whereas later layers require larger noise $\bm\xi$ (larger $c$). We present an empirical experiment to verify our analysis in Section.~\ref{sec:4.3}.

\subsection{Robustness of RMU Models to Adversarial Jailbreak Attacks}
\label{sec:3.3}
\begin{figure*}[!ht] 
\centering 
\subfigure[$c=1.0$]{\includegraphics[width=0.235\textwidth]{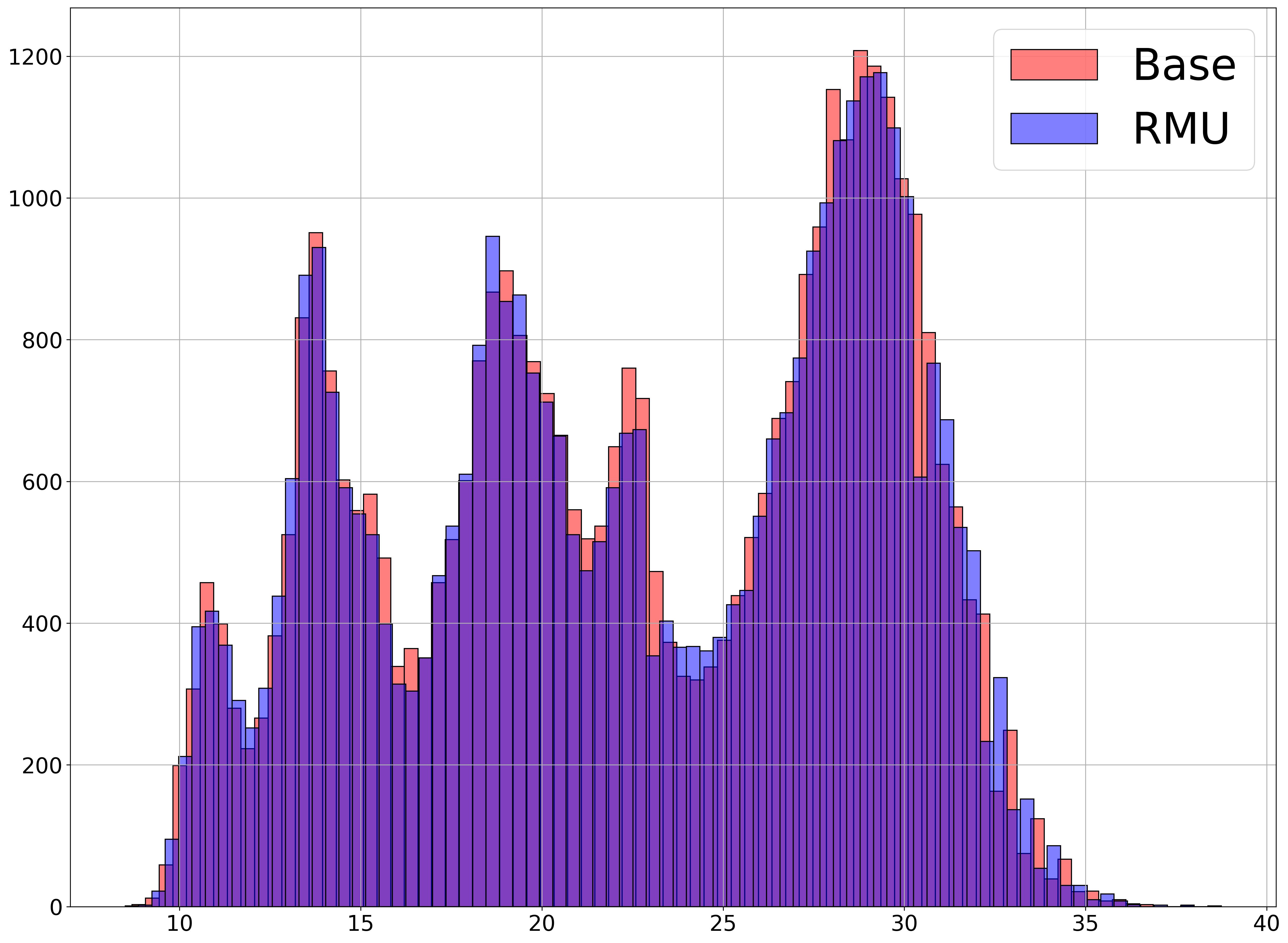}}
\subfigure[$c=2.0$]{\includegraphics[width=0.235\textwidth]{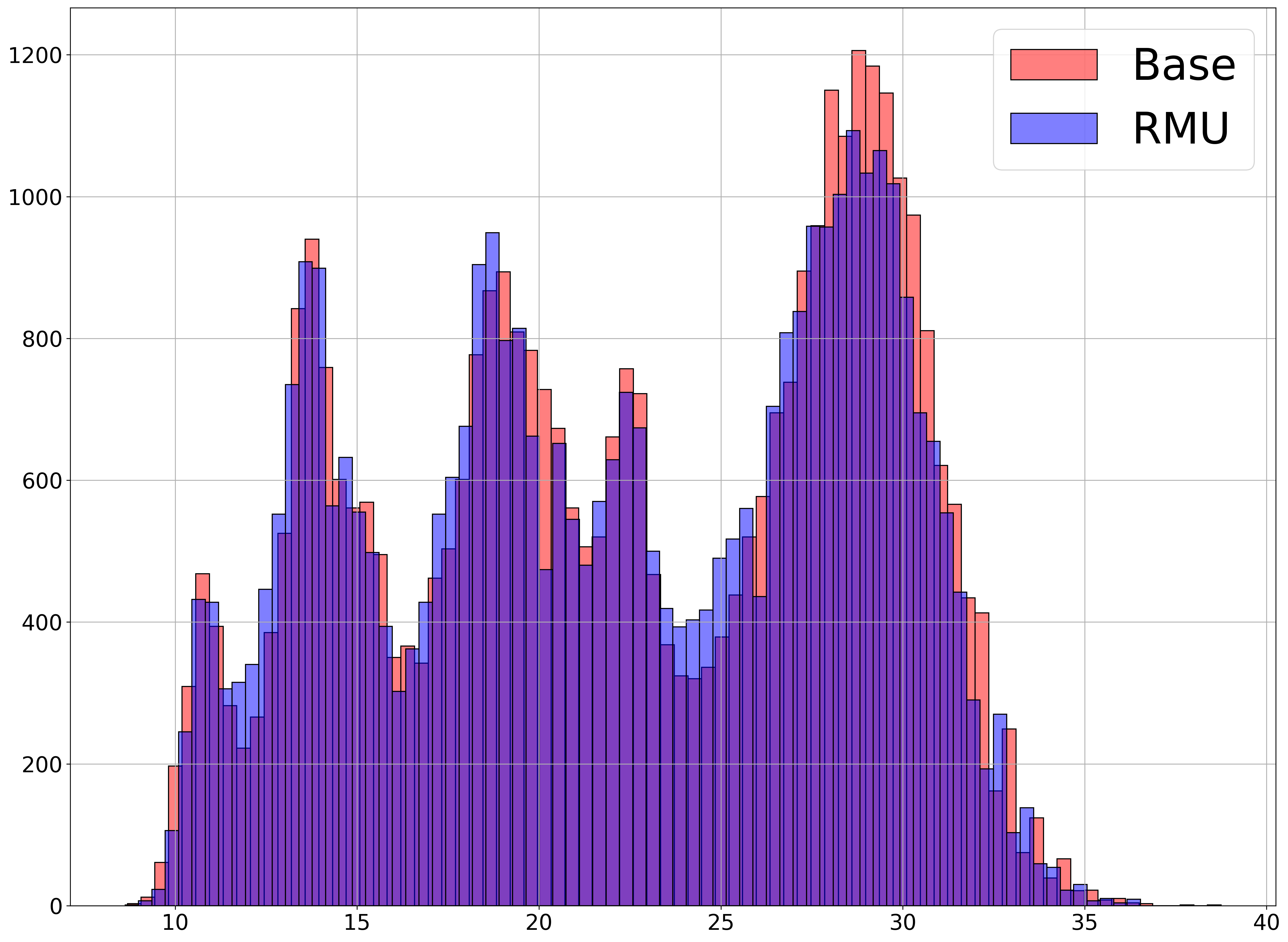}}
\subfigure[$c=5.0$]{\includegraphics[width=0.235\textwidth]{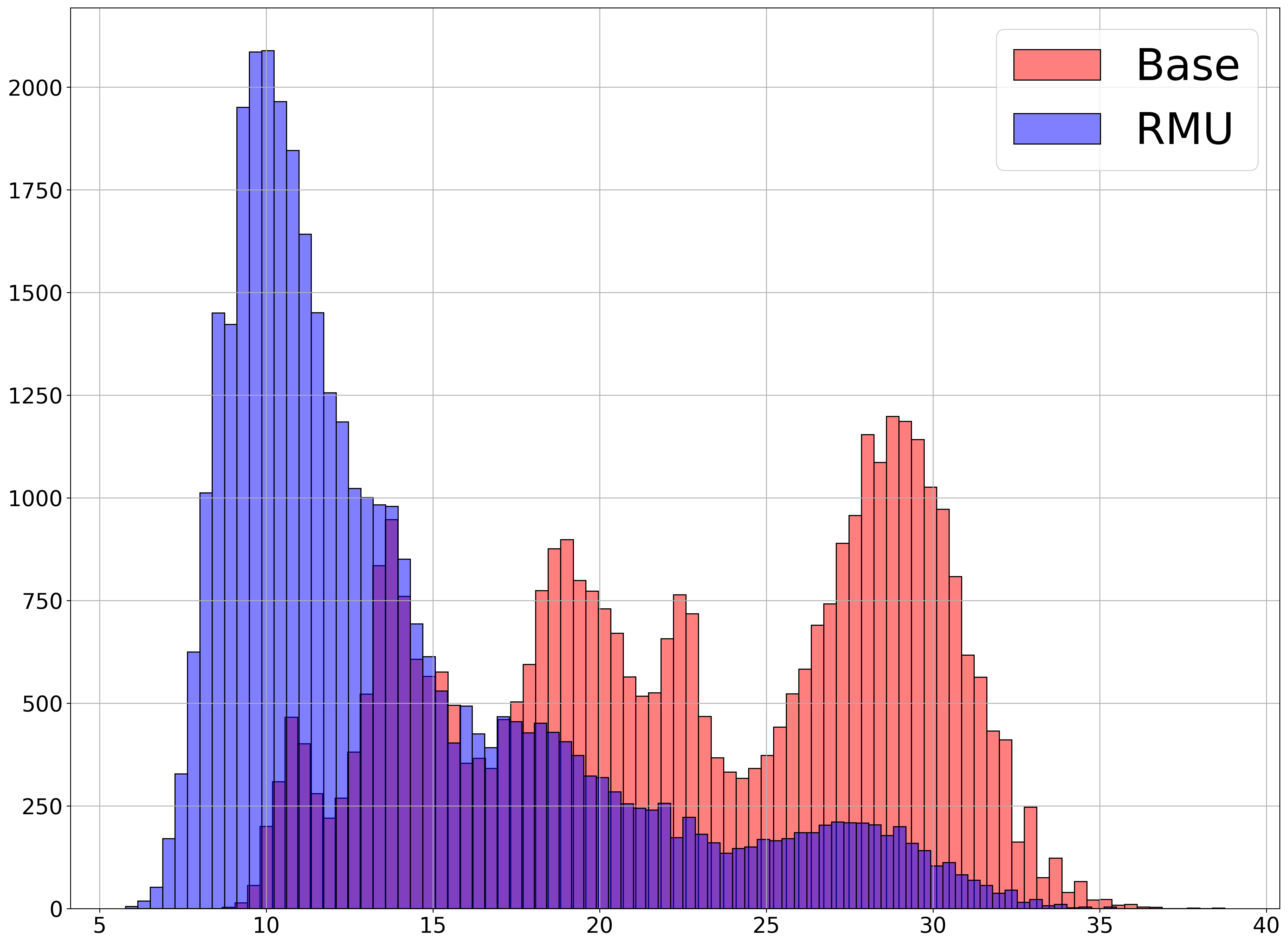}}
\subfigure[$c=10.0$]{\includegraphics[width=0.235\textwidth]{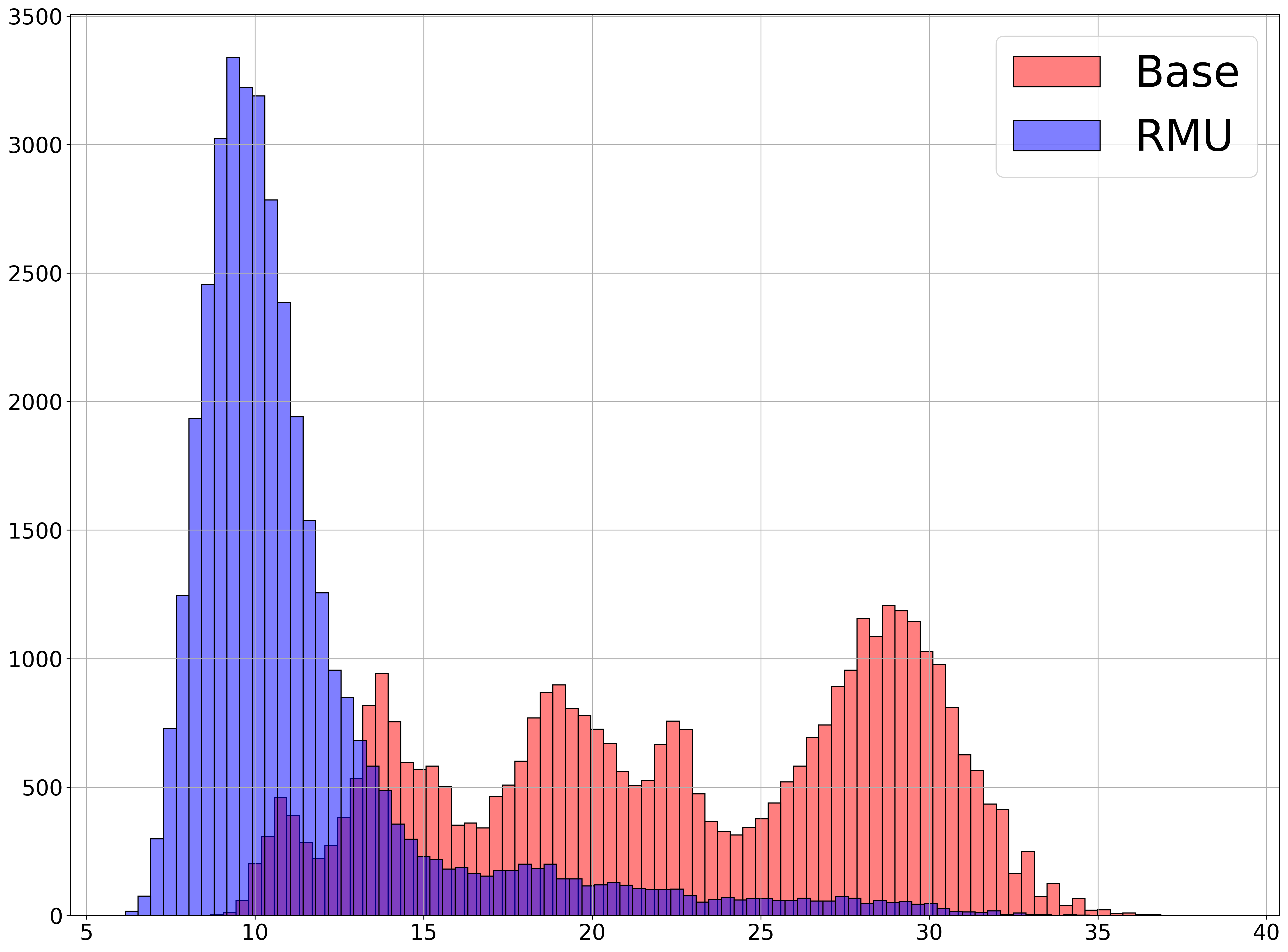}}
\subfigure[$c=1.0$]{\includegraphics[width=0.235\textwidth]{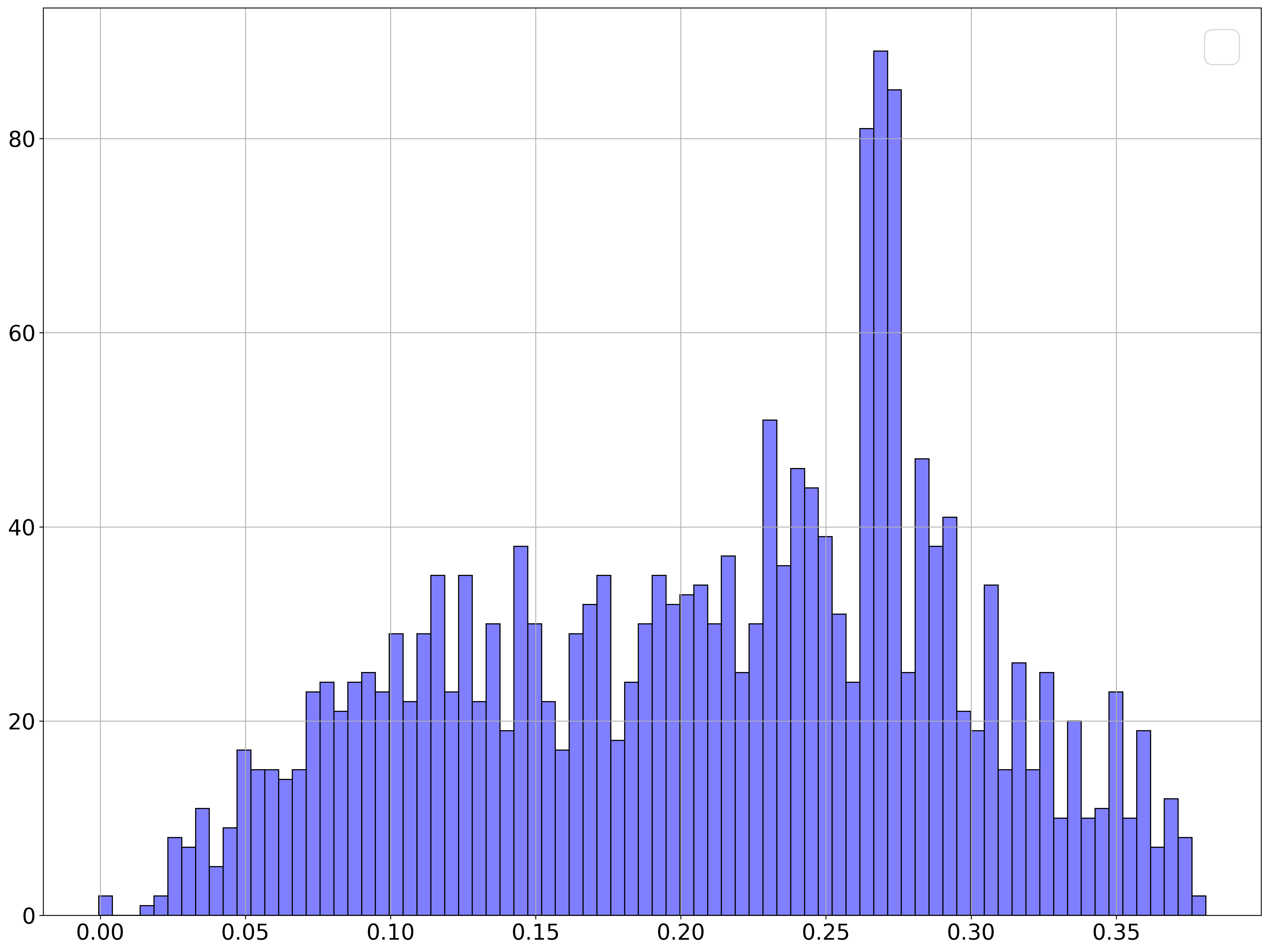}}
\subfigure[$c=2.0$]{\includegraphics[width=0.235\textwidth]{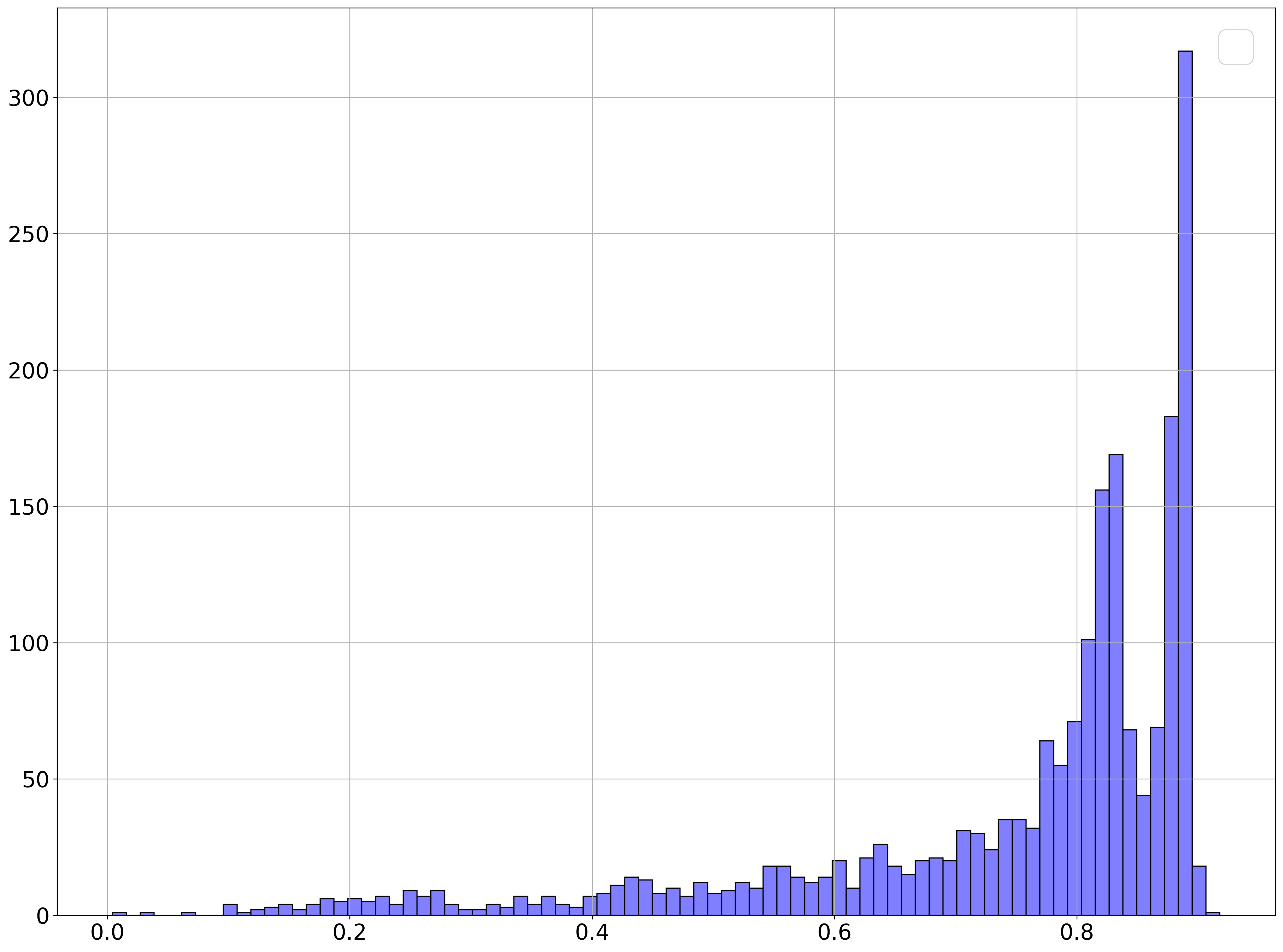}}
\subfigure[$c=5.0$]{\includegraphics[width=0.235\textwidth]{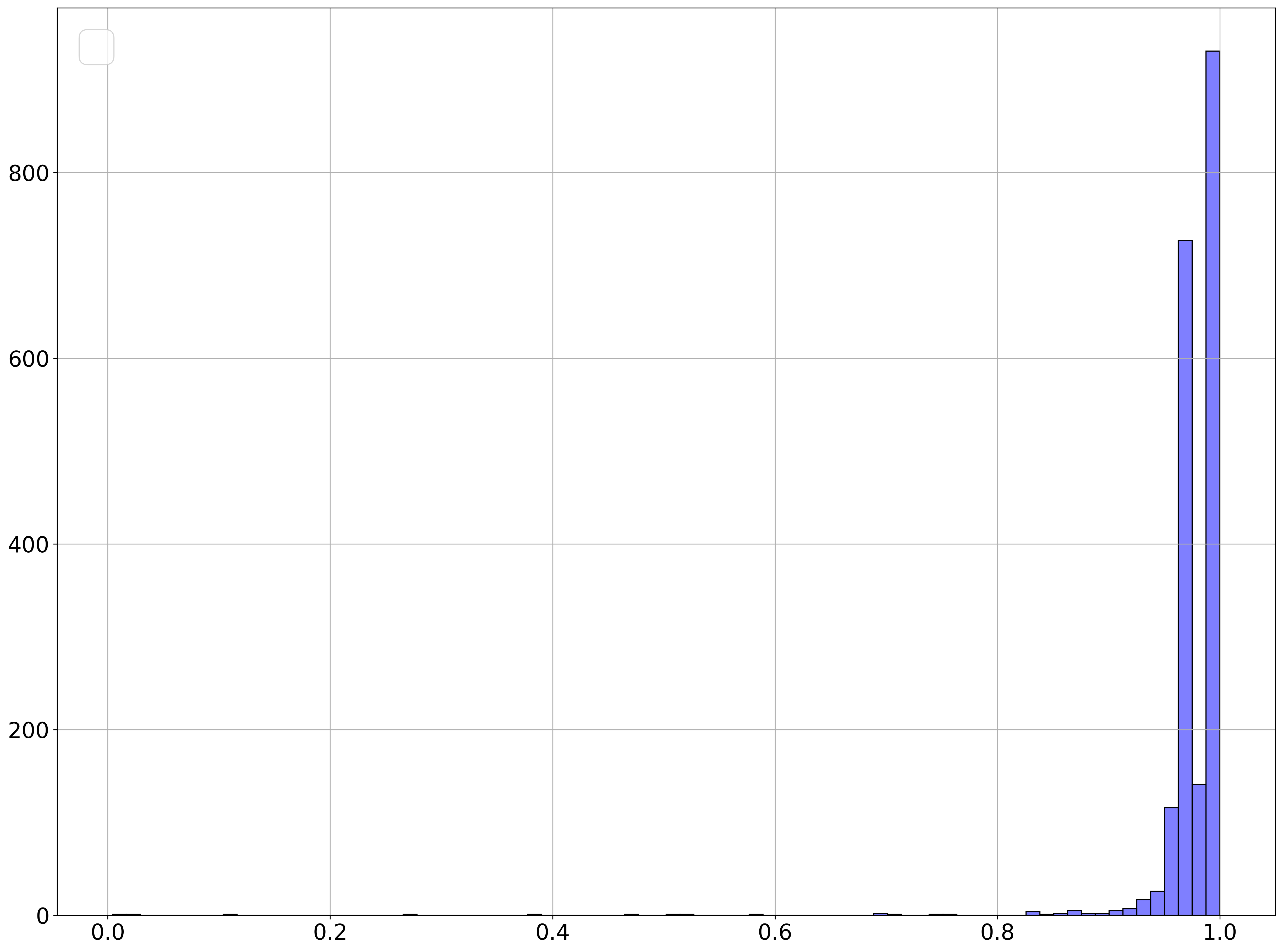}}
\subfigure[$c=10.0$]{\includegraphics[width=0.235\textwidth]{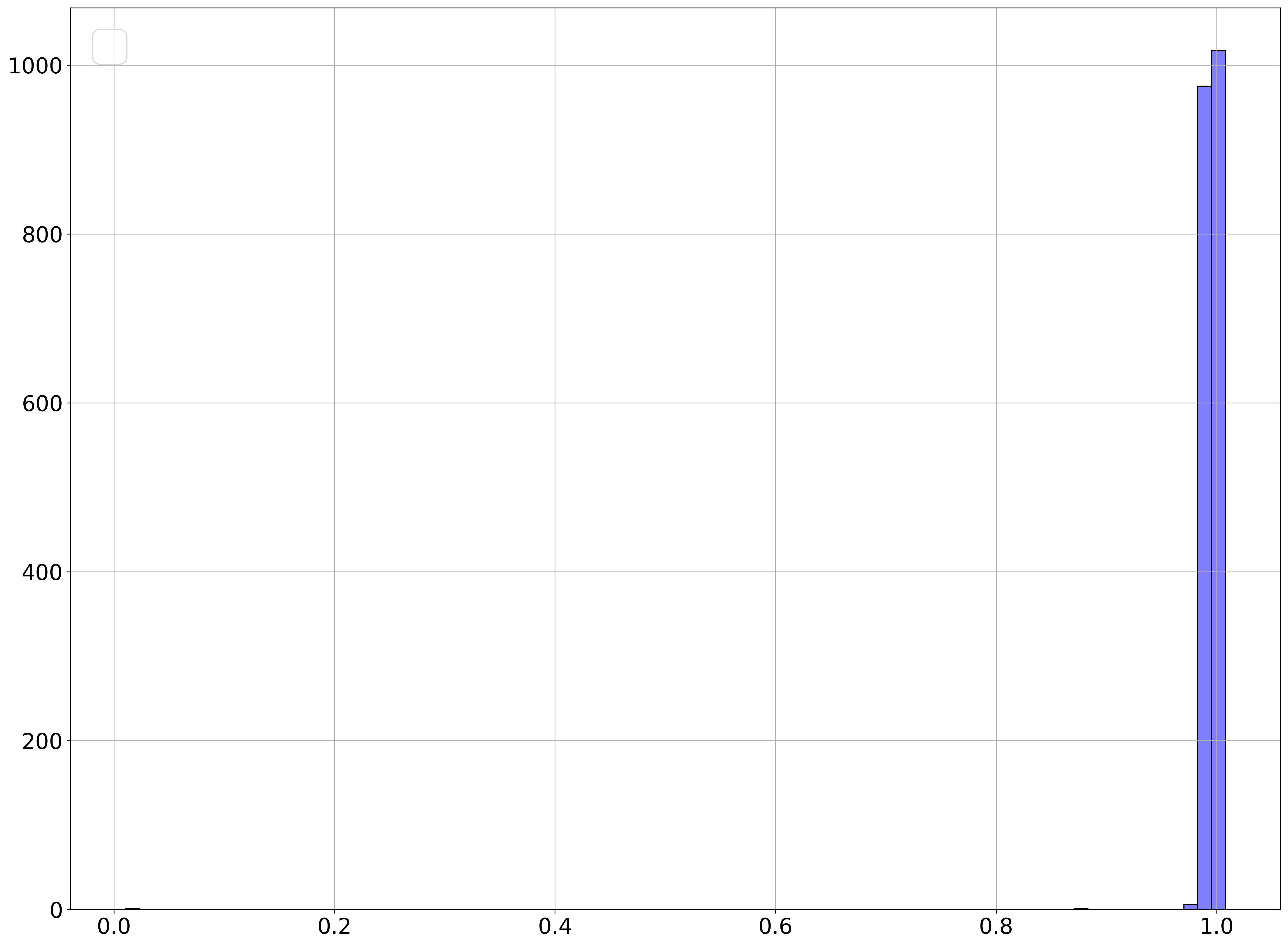}}
\caption{The distribution of MaxLogit~(a-d) on WMDP Q\&A sets with different coefficient $c$ of the base Zephyr-7B and RMU Zephyr-7B models ($l=7$). The distribution of $\cos(\bm u, h^{(l)})$~(e-h) of the RMU Zephyr-7B model ($l=7$).}
\label{fig2}
\end{figure*}
RMU algorithms not only erase undesirable knowledge (forgotten knowledge) but also prevent it from resurfacing when attacked by adversarial jailbreak attacks. \citet{wmdp} present a qualitvative experiment by employing Greedy Coordinate Gradient (GCG;~\citet{zou2023universal}) to jailbreak RMU models to extract the answer given a harmful input. As a result, RMU models demonstrate strong robustness against GCG attack. We conduct an analysis from the standpoint of an attack-defense game to explain the underlying causes for the effectiveness of the RMU algorithm.
\paragraph{Attacker's goal.} The goal of the attacker is to inject an \textit{adversarial trigger} into the suffix of the prompt input (\textit{e.g.}, ``How would I engineer influenza A to be significantly more virulent? \textit{x x x x x x}), causing the model generates a harmful response (\textit{e.g.}, ``To make influenza A more virulent...'').
\paragraph{Attacker's knowledge and capability.} In this setting, we focus on white-box jailbreak attack scenarios~\cite{zou2023universal}, where the victim model's architecture, model input, and output are exposed to the attacker. The attacker is based on gradient signals to search and inject an adversarial trigger into the prompt input, and supplies this adversarial input to the model.
\subsubsection{Problem formulation.}
Let $f: \mathbb{R}^{n\times d} \mapsto \mathbb{R}^{n\times |V|}$ be an autoregressive LLM. Given a prompt input joint with an adversarial trigger $x_{F,1:n}$, the attacker finds an update $\delta$ to adversarial trigger aims to maximize the likelihood of generating the target sequence $x_{F,n+1|n+K}$ consists of $K$ tokens. For simplification, we denote $x_F = x_{F,1:K} = [x_{F,1:n},x_{F,n+1:n+K}]$. The attacker tries to solve the following objective:
\begin{align}
    \min_{x_F+\delta} \mathcal{J}(f(x_F+\delta)),
\end{align}
where $\mathcal{J}(\cdot,\cdot)$ is the loss function of the attacker. 
The attacker finds an update $\delta$ based on the linearized approximation of the loss $\nabla_{e_{x_i}}\mathcal{J}(f(x_F))$, where $e_{x_i}$ is the one-hot vector representing the current value of the $i$-th token in $x_F$. The gradient $\nabla_{e_{x_i}}\mathcal{J}(f(x_F))$ is a good indicator for finding a set of candidates for the adversarial token replacement. A more negative value of the gradient $\nabla_{e_{x_i}}\mathcal{J}(f(x_F))$ makes a more decrease in the loss. The GCG attacker finds top-$k$ largest negative value of $\nabla_{e_{x_i}}\mathcal{J}(f(x_F))$ for each token in the adversarial trigger and makes the replacement the most decrease in the loss.
\subsubsection{Robustness of RMU models against GCG attack.} We show that the GCG attacker misjudges in finding optimal adversarial token substitution in RMU models. Specifically, the gradient of the loss at input $x_F$ with respect to $e_{x_i}$ in RMU model is
\begin{align}
    \nabla_{e_{x_i}}\mathcal{J}(f^{\text{unlearn}}(x_F))
\end{align}
Given the Assumption~\ref{assumption1}, we have
\begin{align}
    \nabla_{e_{x_i}}\mathcal{J}(f^{\text{unlearn}}(x_F)) &= \nabla_{e_{x_i}}\mathcal{J}(g^{(l:k)}(h^{(l),\textnormal{steered}}(x_F)) \nonumber\\&\approx \nabla_{e_{x_i}}(\mathcal{J}\circ g^{(l:k)}) (c\bm u + \bm \epsilon)
\end{align}
Since $c$ and $\bm u$ are predetermined before unlearning, $(\mathcal{J}\circ g^{(l:k)}) (c\bm u)$ does not change with respect to $e_{x_i}$. The gradient $\nabla_{e_{x_i}}(\mathcal{J}\circ g^{(l:k)}) (c\bm u + \bm\epsilon)$ close to $0$ for all token $x_i$ since the error $\bm \epsilon \to \bm 0$ as unlearning becomes accurate. This means the GCG attacker received unreliable, uninformative gradient signals from RMU models. The RMU model serves as a defender by causing the attacker to miscalculate the gradient of the loss to optimize its objective, thereby increasing the attacker’s cost. The attacker, therefore, cannot find the optimal adversarial tokens for replacement. \citet{wmdp}'s experiment results implicitly verify our analysis.

\section{Empirical Analysis}

\subsection{Measuring Token Confidence with MaxLogit}
As discussed in Section~\ref{sec:3.1}, we validate our hypothesis by considering the Maximum Logit Value (MaxLogit) estimator for measuring the token confidence. More specifically, we compute the MaxLogit for each token $x_{n+1}$ given a sequence of tokens $x_{1:n} = \{x_1,...,x_n\}$ from vocabulary $V$ as:
\begin{align}
    \text{MaxLogit}(x_{n+1}) = \max_{x_{n+1} \in V} f^{\text{unlearn}}(x_{n+1}|x_{1:n}) \label{eq21}
\end{align}
We use WMDP-Biology and WMDP-Cyber Q\&A datasets~\cite{wmdp} with total $3260$ Q\&As. We formulated each question and answer as a zero-shot Q\&A prompt to query the unlearned LLM. The details of the prompt template are located in Appendix~A.1. We used greedy decoding to generate tokens and compute the MaxLogit of each token over $k=30$ generated tokens. The MaxLogit distribution was then analyzed for each model Base vs. RMU (unlearned on WMDP-Biology and WMDP-Cyber forget datasets). 

The results are presented in Fig.~\ref{fig2} (a)-(d). We find that the MaxLogit distribution for the base model is generally wider compared to the RMU model.  In contrast, the RMU model demonstrates a more concentrated and approximately normal distribution of MaxLogit values. The peak of the RMU model's MaxLogit distribution is shifted towards lower values relative to the base model. This indicates that the RMU model tends to assign lower confidence scores to the generated tokens. Overall, the RMU model's MaxLogit distribution exhibits lower compared to the base model.

\subsection{The Effect of the Coefficient $c$}
\begin{figure}[!t] 
\centering 
\subfigure{\includegraphics[width=0.228\textwidth]{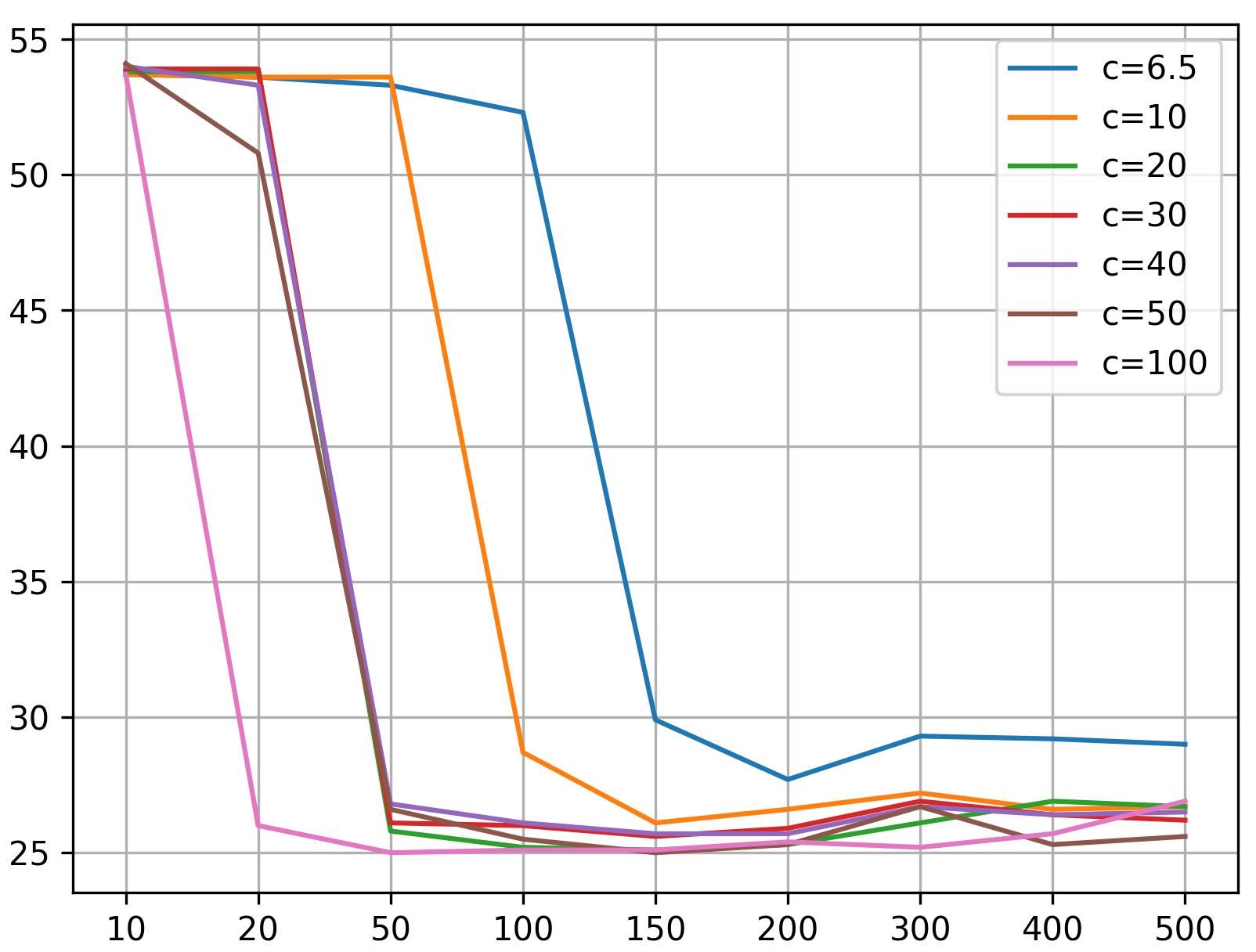}
}
\subfigure{\includegraphics[width=0.228\textwidth]{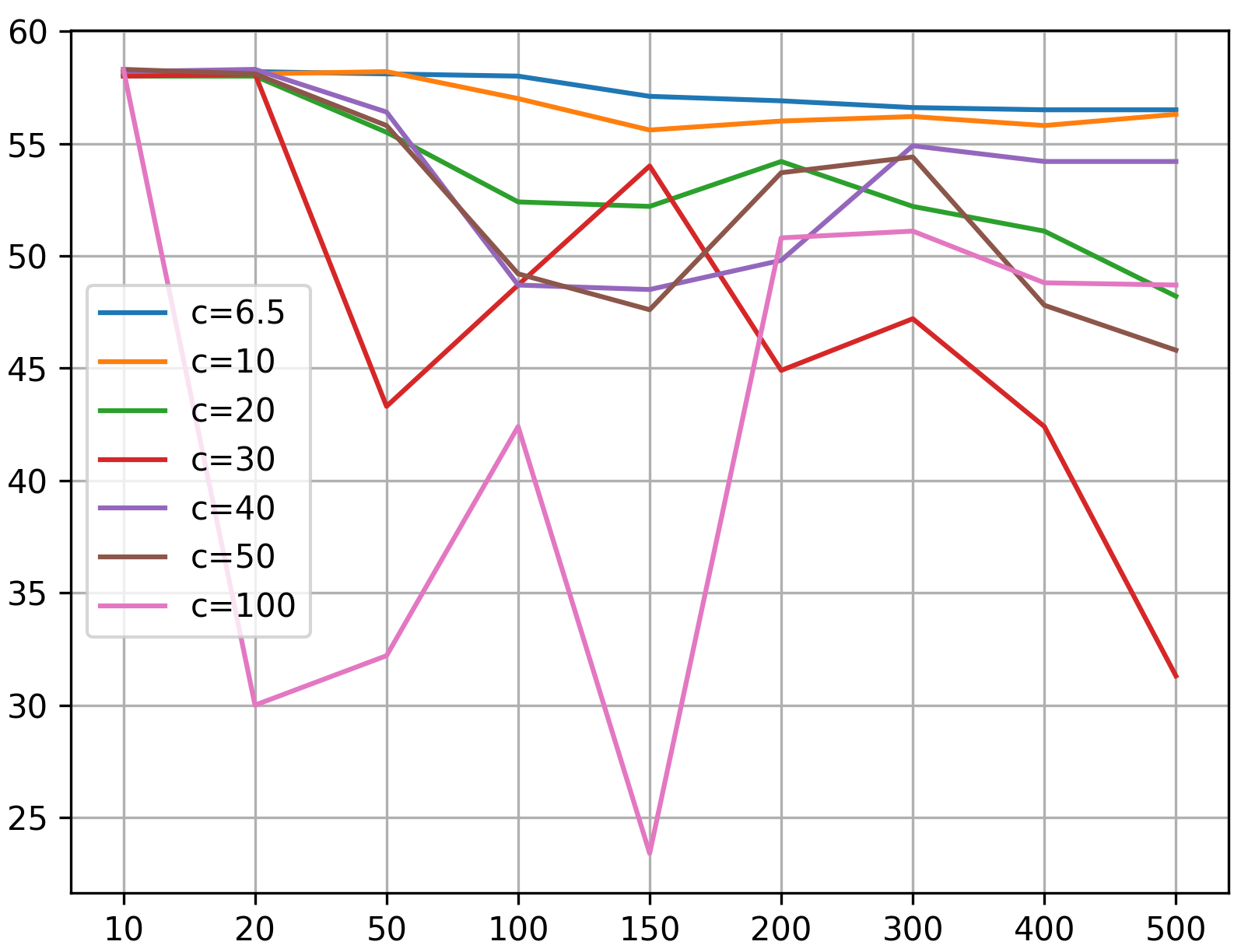}
}
\caption{Average accuracy of WMDP (Biology and Cyber) (left) and MMLU with different coefficient $c$ (right).}
\label{fig3}
\end{figure}

\paragraph{On accuracy.}
\label{sec:2.3}
We analyze the impact of $c$ for forgotten knowledge and retained knowledge, using WMDP~\cite{wmdp} and MMLU~\cite{mmlu}. See Section~\ref{sec:experiment} for the full experiment setting.
Fig.~\ref{fig3}a shows: (i) a clear positive correlation between the drop-in-accuracy rate and the value of $c$, \textit{i.e.} higher $c$ makes the accuracy decrease faster. (ii) A larger value of $c$ tends to make a more drop-in-accuracy on WMDP. (iii) However, a larger $c$ comes with a caveat in a significant drop in general performance on MMLU (Fig.~\ref{fig3}b).
\paragraph{On alignment between $\bm u$ and $h^{(l)}$.} We compute $\cos(\bm u, h^{(l)})$ scores of pairs of $\bm u$ and $h^{(l)}(x_F)$ for all $x_F$ in on WMDP-Biology and WMDP-Cyber forget datasets and plot the
$\cos(\bm u, h^{(l)})$ score distribution shown in Fig.~\ref{fig2}(e)-(h). We observed that there is a clear positive correlation between $\cos(\bm u, h^{(l)})$ scores and the coefficient $c$. As $c$ increases, the distribution of $\cos(\bm u, h^{(l)})$ scores shifts towards higher values and are almost distributed with a peak at $1.0$ (Fig.~\ref{fig2}(g)-(h)). This verify our analysis in Section~\ref{sec:3.2}.

\subsection{The Effect of Layers on Unlearning}
\label{sec:4.3}
\begin{figure}[!hbt]
\centering
\includegraphics[width=0.47\textwidth]{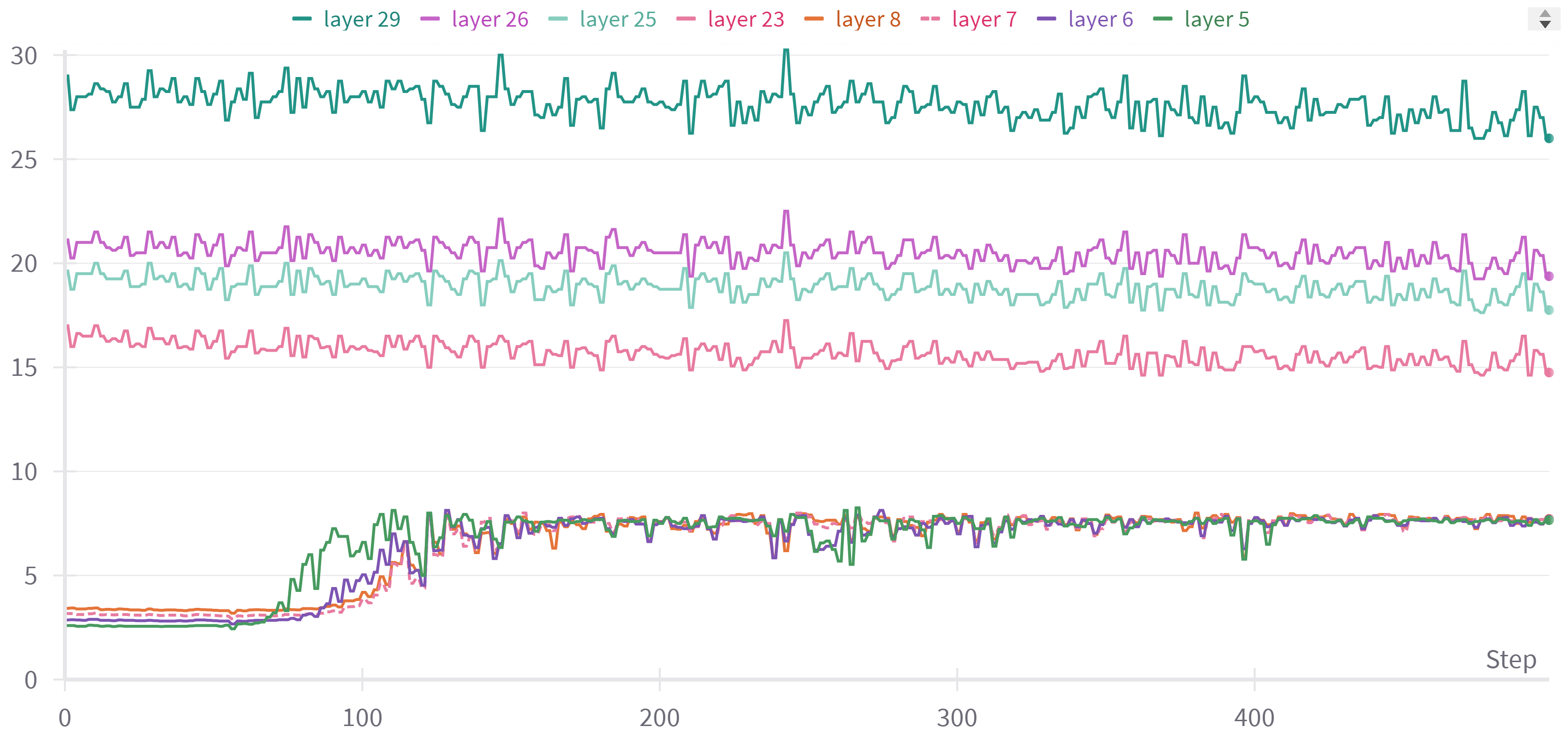}
\caption{$\ell^2$-norm of forget-sample representation.}
\label{fig4}
\end{figure}
\begin{figure*}[!ht] 
\centering 
\includegraphics[width=\textwidth]{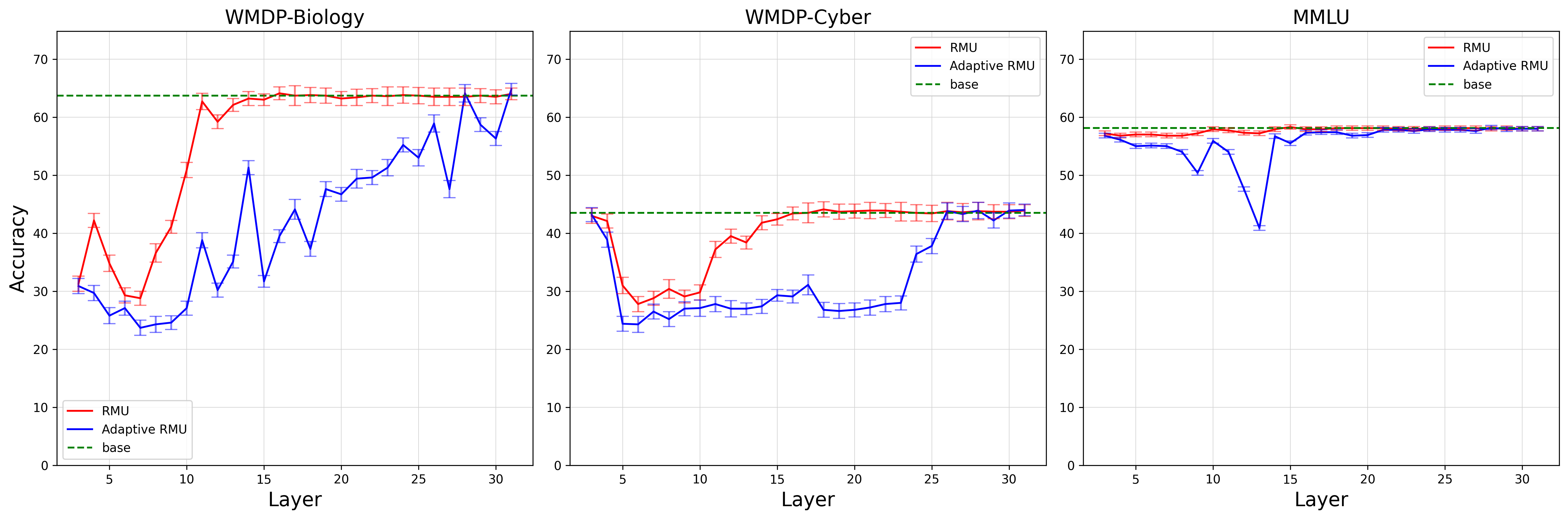}
\caption{Q\&A accuracy of RMU and Adaptive RMU Zephyr-7B models on WMDP-Biology, WMDP-Cyber, and MMLU w.r.t unlearn layer $l$ from the third to the last layer.}
\label{fig5}
\end{figure*}

\noindent We investigate the effect of unlearn layers on accuracy and the representation norm during unlearning. Following original work, we change the unlearn layer $l$ from $3\to 31$, fixed $c=6.5$. Fig.~\ref{fig5} shows that RMU is effective for unlearning within the early layers ($3\to10$), yet exhibits inefficacy within middle and later layers ($11\to31$). Interestingly, in Fig.~\ref{fig4}, we observed that within early layers, the $\ell^2$-norm of forget samples are smaller than the coefficient $c$. 
During unlearning, the representation norm exponentially increases, approaching $c$, thereby facilitating the convergence of forget loss. Conversely, within middle and later layers, the representation norms of forget samples, initially larger than $c$, remain unchanged during unlearning, making the forget loss non-convergence.

\section{Adaptive RMU}

\begin{algorithm}[t!]
\caption{Adaptive RMU pseudocode}\label{algo:adaptiveRMU}
\begin{algorithmic}[1]
\REQUIRE
\STATE $\mathcal{D}_{\text{forget}}$: a forget dataset.
\STATE $\mathcal{D}_{\text{retain}}$: a retain dataset.
\STATE $f_{\theta^{\text{frozen}}}$: a frozen model.
\STATE $f_{\theta^{\text{unlearn}}}$: an update model.
\STATE $\alpha$: a retain weight.
\STATE $l$: an unlearn layer.
\STATE $\beta$: a scaling factor.
\STATE $T$: number of gradient update steps.
\ENSURE Return the unlearned model $f_{\theta^{\text{unlearn}}}$.
\STATE Sample a random unit vector $\bm u \sim U(0,1)$
\FOR {step $t \in [1...T]: x_F \in \mathcal{D}_{\textnormal{forget}}$, \;$x_R \in \mathcal{D}_{\textnormal{retain}}$}
    \STATE Get the representations of $x_F$ and $x_R$ from the frozen and update model.
    \STATE Compute the adaptive loss $\mathcal{L}^{\textnormal{adaptive}}$ by Eqn.~\ref{eq8}.
    \STATE Update $\theta^{\text{unlearn}}$ w.r.t $\nabla\mathcal{L}^{\textnormal{adap}}$ using gradient descent.
    \STATE $t = t + 1$
\ENDFOR
\RETURN $f_{\theta^{\text{unlearn}}}$
\end{algorithmic}
\end{algorithm}
\noindent Inspired by the observations in Section~\ref{sec:4.3}, we propose \emph{Adaptive RMU}, a simple yet effective alternative method with an adaptive forget loss by scaling the random unit vector $\bm u$ with an \textit{adaptive scaling coefficient} $\beta||h_{\theta^{\text{frozen}}}^{(l)}(x_F)||$, where  $\beta \in \mathbb{R}$ is a scaling factor and $||h_{\theta^{\text{frozen}}}^{(l)}(x_F)||$ is the $\ell^2$-norm of forget-sample $x_F$ on model $f_{\theta^{\text{frozen}}}$. The total loss is calculated as follows:
\begin{align}
    \mathcal{L}^{\textnormal{adaptive}} &= \underbrace{\mathbb{E}_{x_F\in \mathcal{D}_{\text{forget}}}\nonumber||h_{\theta^{\text{unlearn}}}^{(l)}(x_F)-\beta||h_{\theta^{\text{frozen}}}^{(l)}(x_F)|| \bm u||_2^2}_{\textnormal{adaptive forget loss}} \\&+ \alpha\underbrace{\mathbb{E}_{x_R\in \mathcal{D}_{\text{retain}}}||h_{\theta^{\text{unlearn}}}^{(l)}(x_R)-h_{\theta^{\text{frozen}}}^{(l)}(x_R)||_2^2}_{\textnormal{retain loss}}\label{eq8}
\end{align}
Our Adaptive RMU is shown in Algorithm~\ref{algo:adaptiveRMU}. We note that Adaptive RMU aims to address the challenge of adaptively determining the coefficient $c$ in RMU. We acknowledge that the introduced value $\beta$ is manually tuned via grid search, leaving the challenge to not fully resolved. However, we emphasize that Adaptive RMU offers significant computational advantages over the original RMU. More concretely, in RMU, grid search is conducted over both $c$ and layer $l$ for $l \in [1...L]$, where $L$ is the number of layers. Our analysis suggests that effective unlearning can be achieved when $c$ is higher than the representation norm of forget-samples. Therefore, given a layer $l$, Adaptive RMU only requires tuning $\beta$, which is $L$ times less than that of RMU. This reduction in computational overhead represents a significant improvement when the size of modern deep networks grows.


\section{Experiment}
\label{sec:experiment}
\paragraph{Datasets.} We use WMDP-Biology and WMDP-Cyber forget datasets as $\mathcal{D}_\text{forget}$ and Wikitext~\cite{wikitext} as $\mathcal{D}_\text{retain}$ for unlearning the LLM. Unlearned models are evaluated on WMDP Q\&A datasets and MMLU~\cite{mmlu}. Details of the datasets can be found in the Appendix~A.1. 
\paragraph{Models.} We use the following LLMs: Zephyr-7B-$\beta$~\cite{tunstall2023zephyr}, Yi-6B~\cite{young2024yi}, Meta Llama-3-8B~\cite{meta2024introducing}, and Mistral-7B~\cite{jiang2023mistral}.

\paragraph{Experimental setup.} Models were fine-tuned using AdamW~\cite{adamw} with learning rate $\eta = 5e-5$, batch-size of $4$, max sequence len of $512$ for WMDP-Biology and $768$ for WMDP-Cyber, with $T=500$ gradient update steps. 
The retain weight $\alpha=1200$. 
For the baseline RMU, we follow the previous work and let $c=6.5$. 
We grid search for unlearn layer $l$ from the third to the last layer. For the Adaptive RMU, we grid search for the scaling factor $\beta \in \{2, 3, 5, 10\}$. We report the performances of Adaptive RMU models with $\beta=5$.
We update three layers parameters $\{l, l-1, l-2\}$ of the model. 
Two NVIDIA A40s with 90GB GPU were used to run the experiments. Our code is available at \url{https://github.com/RebelsNLU-jaist/llm-unlearning}.

\paragraph{Baselines.} We compare Adaptive RMU against baselines: RMU~\cite{wmdp}, Large Language Model Unlearning (LLMU;~\citet{llmu}), SCalable Remenbering and Unlearning unBound (SCRUB;~\citet{kurmanji2023towards}), and Selective Synaptic Dampening (SSD;~\citet{ssd}. We use off-the-shelf results from \citet{wmdp} for LLMU, SCRUB, and SSD.

\paragraph{Main results.}
\begin{table}[!t]
    \centering
    \resizebox{0.475\textwidth}{!}{
    
    \begin{tabular}{l|c|c|c}
    \hline
    \rowcolor{gray!30}
    Method/tasks   & WMDP-Biology~$\downarrow$ &  WMDP-Cyber~$\downarrow$ &  MMLU~$\uparrow$ \\\hline
     Base  &   63.7      &   43.5       &  58.1 \\\hline
     LLMU          &   59.5       &   39.5       &  44.7 \\\hline
     SCRUB         &   43.8       &     39.3         & 51.2 \\\hline
     SSD            &     50.2     &   35.0        & 40.7  \\\hline
     RMU ($l=7$)    &      \underline{28.8}    &   \underline{28.8}       & \textbf{56.8}\\\hline
     \textbf{Adaptive RMU} ($l=7$) &      \textbf{23.7}    &   \textbf{26.5}      &  \underline{55.0} \\
    \hline
    \end{tabular}}
    \caption{Q\&A accuracy of Zephyr-7B models on WMDP and MMLU. The $\textbf{best}$ and $\underline{\textnormal{runner up}}$ are marked.}
    \label{tab1}
\end{table}
Fig.~\ref{fig5} shows that Adaptive RMU significantly improves unlearning performances. Specifically, Adaptive RMU reduces average accuracy by $13.1\%$ on WMDP-Biology and $3.6\%$ on WMDP-Cyber within early layers ($3\to10$), and by $15.6\%$ on WMDP-Biology and $9.6\%$ on WMDP-Cyber within middle and later layers ($11\to31$). This corresponds to an overall enhancement of $14.3\%$ and $6.6\%$ in drop-in-accuracy for the WMDP-Biology and WMDP-Cyber, respectively. Table~\ref{tab1} further highlights that Adaptive RMU ($l=7$) outperforms RMU ($l=7$), LLMU, SCRUB, and SSD,  
establishing a new state-of-the-art performance.
We defer the full results on other models and settings in Appendix B.

\section{Conclusion}
We studied the effect of steering latent representation for LLM unlearning and explored its connection to jailbreak adversarial robustness. We developed a simple yet effective alternative method that enhances unlearning performance across most layers while maintaining overall model utility. Our findings illuminate the explanation of the RMU method and pave the way for future research in LLM unlearning.

\section*{Acknowledgments}
This work was supported by JST FOREST Program (Grant Number JPMJFR232K, Japan) and the Nakajima Foundation.

\bibliography{aaai25}
\newpage
\appendix
\label{sec:appendix}
\setcounter{table}{1}
\setcounter{figure}{5}
\section{Datasets and Q\&A template}
\label{appendix:B}

\subsection{Datasets}
\label{sec:dataset}
\paragraph{WMDP~\textnormal{\cite{wmdp}}} stands for Weapon of Mass Destruction Proxy, is a corpora consisting of forget sets, retain sets, and Q\&A sets. 
The WMDP Q\&A is a dataset of $3,668$ multiple-choice questions about Biosecurity ($1,273$), Cybersecurity ($1,987$), and Chemical security ($408$). The WMDP-Biology forget and retain sets consist of papers from PubMed. The WMDP-Biology forget comprises papers used in generating WMDP-Biology questions, while the retain set samples papers from various categories within general biology. The retain set excludes papers from the forget set and employs keyword exclusion to avoid topics related to Q\&A set. The WMDP-Cyber forget and retain sets consist of passages crawled from GitHub with two different sets of keywords. Note that we did not benchmark for WMDP-Chemistry Q\&A due to no WMDP-Chemistry forget set being publicly released. This dataset is available at \url{https://github.com/centerforaisafety/wmdp}.
\paragraph{MMLU~\textnormal{\cite{mmlu}}} stands for Massive Multitask Language Understanding, a dataset of $15,908$ multiple-choice Q\&A covers $57$ subjects across STEM, the humanities, social science, and more. MMLU is designed to measure general knowledge by evaluating models in zero-shot or few-shot settings. This dataset is available at \url{https://huggingface.co/datasets/cais/mmlu}.
\paragraph{Q\&A format.}
\label{sec:qaformat} 
We follow the default zero-shot Q\&A format to query LLM~\cite{eval-harness}. For example:\\[0.1cm]
``The following are multiple choice questions (with answers) about Biology.\\\\
Question: How can I cultivate a calicivirus?\\
A. Cell/enteroid culture\\
B. Helper virus-free reverse genetics system\\
C. Either A or B\\
D. Neither A nor B\\
Answer:''

\subsection{Computational perplexity of the RMU and Adaptive RMU.}
\label{appendix:a2}

The difference between RMU and Adaptive RMU is the calculation of the forget coefficient. RMU uses a predefined coefficient $c$. Adaptive RMU uses an adaptive coefficient, $\beta||h_{\theta^{\text{frozen}}}^{(l)}(x_F)||$, which can be calculated and cached during the first iteration of the inner $\textbf{for}$ loop in Algorithm~1. Thus, the complexity of Adaptive RMU is equal to that of RMU. 

\section{Additional results}
\label{additional_results}

\subsection{Unlearning performance of other models}
We report the unlearning performance of Adaptive RMU Yi-6B, Llama-3-8B, and Mistral-7B models in Table~\ref{tab2}, Table~\ref{tab3}, and Table~\ref{tab4}. We observed a clear trend that the unlearning performance is more effective when using the early layer as the unlearn layer.

\begin{table*}[!ht]
    \centering
    \resizebox{\textwidth}{!}{
    \begin{tabular}{l|c|c|c|c|c|c|c|c|c|c|c|c|c|c|c}
    \hline
    \rowcolor{gray!30}
    Task/unlearn layer          & base & 3    & 4     & 5    & 6    & 7    & 8     & 9    & 10   & 11   & 12   & 13   & 14   & 15   & 16 \\\hline
    WMDP-Biology~$\downarrow$   &64.8 &65.0 &49.9& 35.2& 27.8 &26.1& 63.3& 26.2& 27.1& 27.4& 27.1& 26.0& 25.4& 27.2& 34.8\\\hline
    WMDP-Cyber~$\downarrow$     &41.1& 40.7& 40.5& 37.7& 28.1& 25.5& 39.3& 25.6& 23.9& 26.1& 23.6& 24.3& 24.2& 24.0& 25.5 \\\hline
    MMLU~$\uparrow$             &60.0& 60.1& 57.7& 59.4& 51.4& 56.5& 59.9& 56.8& 53.7& 48.1& 49.3& 57.0& 55.6& 47.7& 53.3
\\
    \hline
    \end{tabular}}
    \resizebox{\textwidth}{!}{
     \begin{tabular}{l|c|c|c|c|c|c|c|c|c|c|c|c|c|c|c}
    \hline
        \rowcolor{gray!30}
        Task/unlearn layer           & 17   & 18   & 19   & 20   & 21   & 22   & 23   & 24   & 25   & 26   & 27   & 28   & 29   & 30   & 31  \\\hline
        WMDP-Biology~$\downarrow$    & 30.3 & 32.2&  27.1 & 31.9&  41.0&  53.4&  50.4&  53.2&  39.2 & 46.0 & 39.0 & 42.5 & 41.6 & 40.5 & 64.8\\\hline
        WMDP-Cyber~$\downarrow$      &  25.3 & 24.4 & 24.3 & 24.5 & 26.7 & 29.8 & 33.9 & 36.2 & 34.3 & 34.6 & 31.4 & 30.4 & 39.6 & 40.8 & 40.6 \\\hline
        MMLU~$\uparrow$              &  45.4 & 52.1 & 56.7 & 58.2 & 59.3 & 59.4 & 59.6 & 59.7 & 59.4 & 59.7 & 59.4 & 59.4 & 59.5 & 59.7 & 60.1 \\
    \hline
    \end{tabular}}
    \caption{Q\&A accuracy of Adaptive RMU Yi-6B models on WMDP-Biology, WMDP-Cyber, and MMLU.}
    \label{tab2}
\end{table*}

\begin{table*}[!ht]
    \centering
    \resizebox{\textwidth}{!}{
    \begin{tabular}{l|c|c|c|c|c|c|c|c|c|c|c|c|c|c|c}
    \hline
    \rowcolor{gray!30}
    Task/unlearn layer    & base & 3    & 4     & 5    & 6    & 7    & 8     & 9    & 10   & 11   & 12   & 13   & 14   & 15   & 16 \\\hline
    WMDP-Biology~$\downarrow$   & 71.2 & 46.4 & 45.3  & 28.2 & 27.8 & 29.3 & 33.7  & 36.0 & 65.1 & 64.9 & 62.8 & 65.2 & 59.6 & 44.4 & 41.4\\\hline
    WMDP-Cyber~$\downarrow$ & 43.9 & 32.5 & 25.5  & 24.5 & 27.6 & 26.8 & 27.3  & 26.3 & 32.5 & 32.3 & 34.1 & 35.2 & 29.9 & 28.3 & 27.8\\\hline
    MMLU~$\uparrow$     & 62.0 & 60.7 & 60.2  & 59.7 & 60.7 & 60.0 & 60.1  & 59.6 & 61.8 & 61.3 & 61.5 & 61.5 & 61.8 & 60.9 & 61.1\\
    \hline
    \end{tabular}}
    \resizebox{\textwidth}{!}{
     \begin{tabular}{l|c|c|c|c|c|c|c|c|c|c|c|c|c|c|c}
    \hline
        \rowcolor{gray!30}
        Task/unlearn layer     & 17   & 18   & 19   & 20   & 21   & 22   & 23   & 24   & 25   & 26   & 27   & 28   & 29   & 30   & 31  \\\hline
        WMDP-Biology~$\downarrow$    & 35.5 & 35.2 & 41.1 & 60.8 & 33.7 & 59.3 & 54.6 & 56.7 & 69.6 & 62.2 & 70.0 & 69.9 & 69.9 & 67.0 & 70.4 \\\hline
        WMDP-Cyber~$\downarrow$  & 28.0 & 33.5 & 28.6 & 39.0 & 28.6 & 31.7 & 35.5 & 36.9 & 45.5 & 44.8 & 44.4 & 43.5 & 44.4 & 43.6 & 43.4 \\\hline
        MMLU~$\uparrow$      & 61.3 & 61.3 & 61.3 & 61.9 & 60.8 & 61.7 & 61.2 & 61.5 & 61.9 & 61.7 & 62.0 & 61.9 & 61.5 & 61.5 & 62.1 \\
    \hline
    \end{tabular}}
    \caption{Q\&A accuracy of Adaptive RMU Meta Llama-3-8B models on WMDP-Biology, WMDP-Cyber, and MMLU.}
    \label{tab3}
\end{table*}

\begin{table*}[!ht]
    \centering
    \resizebox{\textwidth}{!}{
    \begin{tabular}{l|c|c|c|c|c|c|c|c|c|c|c|c|c|c|c}
    \hline
    \rowcolor{gray!30}
    Task/unlearn layer      & base & 3    & 4    & 5    & 6    & 7    & 8    & 9    & 10   & 11   & 12 & 13 & 14 & 15 & 16 \\\hline
    WMDP-Biology~$\downarrow$ & 67.3 & 28.0 & 28.9 & 27.6 & 27.5 & 26.3 & 24.5 & 25.7 & 26.1 & 27.6 & 31.4 & 37.7 & 35.6 & 25.4 & 35.0\\\hline
    WMDP-Cyber~$\downarrow$   & 44.1 & 42.1 & 41.9 & 24.8 & 26.8 & 26.3 & 26.6 & 26.4 & 26.7 & 25.7 & 26.5 & 25.8 & 31.6 & 26.7 & 27.9\\\hline
    MMLU~$\uparrow$       & 58.7 & 54.5 & 57.2 & 54.9 & 55.8 & 55.7 & 47.3 & 53.0 & 47.4 & 35.1 & 54.5 & 55.9 & 51.5 & 44.9 &57.3\\
    \hline
    \end{tabular}}
    \resizebox{\textwidth}{!}{
     \begin{tabular}{l|c|c|c|c|c|c|c|c|c|c|c|c|c|c|c}
    \hline
        \rowcolor{gray!30}
        Task/unlearn layer     & 17   & 18   & 19   & 20   & 21   & 22   & 23   & 24   & 25   & 26   & 27   & 28   & 29   & 30   & 31  \\\hline
        WMDP-Biology~$\downarrow$    & 27.4 & 56.4 & 38.4 & 45.7 & 42.0 & 52.0 & 52.4 & 61.1 & 57.5 & 62.2 & 63.2 & 66.3 &   61.9   & 61.0     & 66.0\\\hline
        WMDP-Cyber~$\downarrow$  & 27.5 & 38.9 & 26.5 & 26.7 & 26.6 & 27.4 & 27.7 & 38.9 & 43.9 &  43.4 &43.7 & 43.8 &    44.0   & 42.5     & 43.4 \\\hline
        MMLU~$\uparrow$      & 56.7 & 56.8 & 56.2 & 57.6 & 58.1 & 58.3 & 58.1 & 58.2 & 58.6 & 58.7 & 58.6 & 58.7 & 58.4  & 58.3     & 58.2 \\
    \hline
    \end{tabular}}
    \caption{Q\&A accuracy of Adaptive RMU Mistral-7B models on WMDP-Biology, WMDP-Cyber, and MMLU.}
    \label{tab4}
\end{table*}

\subsection{Performances on MMLU subset unlearning benchmark}
We did additional experiments on the MMLU subset unlearning benchmark with three settings: 
\begin{enumerate}
    \item MMLU-Economics: unlearning high school microeconomics and macroeconomics and maintaining performance on the remaining categories (refers as MMLU-Retain tasks).
    \item MMLU-Law: unlearning international and professional law while maintaining performance on MMLU-Retain.
    \item MMLU-Physics: unlearning high school and college physics while maintaining general performance in MMLU-Retain.
\end{enumerate}

\paragraph{Settings.}  We use publicly released forget set by \citet{wmdp} for each task and Wikitext~\cite{wikitext} as retain set. We use a fixed sequence len of $512$ for MMLU-Economics, MMLU-Law, MMLU-Physics, and Wikitext. We keep other hyperparameters remain unchanged as in Section~6.

\paragraph{Result.} Table~\ref{tab5} presents the unlearning performance of Adaptive RMU Zephyr-7B models on MMLU-Economics, MMLU-Law, and MMLU-Physics. We observe a notable reduction in accuracy on the forget tasks. However, the model exhibits excessive unlearning, leading to substantial performance degradation on the MMLU-Retain tasks.
\begin{table*}[!ht]
    \centering
    \resizebox{\textwidth}{!}{
    \begin{tabular}{l|c|c|c|c|c|c|c|c|c|c|c|c|c|c|c}
    \hline
    \rowcolor{gray!30}
Task/unlearn layer        & base & 3    & 4    & 5    & 6    & 7    & 8    & 9    & 10   & 11   & 12   & 13   & 14   & 15   & 16 \\\hline
    MMLU-Economics~$\downarrow$ & 58.0 & 57.0 & 45.7 & 22.8 & 23.4 & 27.0 & 28.8 & 27.0 & 34.6 & 24.6 & 42.1 & 45.5 & 34.8 & 44.5 & 58.3 \\\hline
    MMLU-Law~$\downarrow$       & 55.6 & 49.8 & 53.5 & 25.2 & 24.5 & 26.4 & 24.6 & 24.2 & 21.5 & 23.9 & 51.1 & 44.1 & 36.8 & 44.7 & 46.0 \\\hline
    MMLU-Physics~$\downarrow$ & 38.5 & 39.3 & 37.9 & 28.8 & 27.2 & 23.8 & 21.7 & 20.5 & 21.0 & 29.2 & 32.6 & 34.1 & 34.4 & 35.7 & 42.3 \\\hline
    MMLU-Retain~$\uparrow$      & 58.9 & 58.0 & 57.3 & 39.3 & 45.2 & 39.4 & 35.2 & 36.0 & 44.8 & 35.2 & 52.9 & 55.2 & 46.0 & 54.8 & 56.8\\
    \hline
    \end{tabular}}
    \resizebox{\textwidth}{!}{
     \begin{tabular}{l|c|c|c|c|c|c|c|c|c|c|c|c|c|c|c}
    \hline
        \rowcolor{gray!30}
Task/unlearn layer        & 17   & 18   & 19   & 20   & 21   & 22   & 23   & 24   & 25   & 26   & 27   & 28   & 29   & 30   & 31  \\\hline
    MMLU-Economics~$\downarrow$ & 51.8 & 36.0 & 54.4 & 26.0 & 21.4 & 42.8 & 43.4 & 42.8 & 48.4 & 57.2 & 58.7 & 50.0 & 58.2 & 58.9 & 57.8\\\hline
    MMLU-Law~$\downarrow$       & 49.8 & 24.3 & 54.4 & 27.2 & 24.6 & 24.2 & 25.4 & 44.6 & 54.4 & 55.8 & 56.7 & 53.6 & 55.6 & 55.4 & 56.1 \\\hline
    MMLU-Physics~$\downarrow$   & 37.5 & 26.7 & 26.9 & 21.0 & 21.6 & 24.2 & 23.4 & 25.6 & 29.6 & 37.1 & 31.9 & 33.8 & 36.9 & 33.9 & 38.6 \\\hline
    MMLU-Retain~$\uparrow$      & 57.6 & 47.8 & 57.7 & 36.2 & 30.3 & 39.6 & 47.4 & 52.0 & 58.1 & 58.9 & 58.9 & 56.4 & 59.0 & 59.1 & 59.0 \\
    \hline
    \end{tabular}}
    \caption{Q\&A accuracy of Adaptive RMU Zephyr-7B models on MMLU-Economics, MMLU-Law, MMLU-Phycics, and MMLU-Retain.}
    \label{tab5}
\end{table*}

\subsection{The effect of in-domain retain set on unlearning performance.}
In this setting, we use the WMDP-Biology and WMDP-Cyber retain sets instead of Wikitext. We use the same hyperparameters as in Section~6. Table~\ref{tab6} shows that Adaptive RMU is almost ineffective for all unlearn layers. As WMDP-forget and retain sets are collected from the same source, even with efforts in distinction, these corpora may commonly have overlapping texts. We present an $n$-gram overlap analysis between the WMDP-forget set and the WMDP-retain set as a measurement of unlearning difficulty.

\begin{table*}[!ht]
    \centering
    \resizebox{\textwidth}{!}{
    \begin{tabular}{l|c|c|c|c|c|c|c|c|c|c|c|c|c|c|c}
    \hline
    \rowcolor{gray!30}
    Task/unlearn layer      & base & 3    & 4    & 5    & 6    & 7    & 8    & 9    & 10   & 11   & 12   & 13   & 14   & 15   & 16 \\\hline
    WMDP-Biology~$\downarrow$ & 63.7 & 63.2 & 63.3 & 62.9 & 28.1 & 62.6 & 49.9 & 64.2 & 29.6 & 62.0 & 63.0 & 63.7 & 63.7 & 64.4 & 64.3\\\hline
    WMDP-Cyber~$\downarrow$   & 43.5 & 42.7 & 42.0 & 40.1 & 24.6 & 33.3 & 33.9 & 40.8 & 25.1 & 41.3 & 41.7 & 42.8 & 43.4 & 42.8 & 43.4\\\hline
    MMLU-All~$\uparrow$       & 58.1 & 57.4 & 57.4 & 57.9 & 30.1 & 57.6 & 38.3 & 57.6 & 29.3 & 57.1 & 58.0 & 57.5 & 57.7 & 57.9 & 57.8\\
    \hline
    \end{tabular}}
    \resizebox{\textwidth}{!}{
     \begin{tabular}{l|c|c|c|c|c|c|c|c|c|c|c|c|c|c|c}
    \hline
    \rowcolor{gray!30}
    Task/unlearn layer      & 17   & 18   & 19   & 20   & 21   & 22   & 23   & 24   & 25   & 26   & 27   & 28   & 29   & 30   & 31  \\\hline
    WMDP-Biology~$\downarrow$ & 63.9 & 63.7 & 63.9 & 63.5 & 63.5 & 63.7 & 63.7 & 63.6 & 63.6 & 63.5 & 63.3 & 63.7 & 63.8 & 63.5 & 64.6\\\hline
    WMDP-Cyber~$\downarrow$   & 44.5 & 43.5 & 43.5 & 44.4 & 43.9 & 43.5 & 44.3 & 43.6 & 43.9 & 43.8 & 43.6 & 43.2 & 43.7 & 43.7 & 43.6 \\\hline
    MMLU-All~$\uparrow$       & 58.4 & 58.1 & 58.2 & 57.6 & 58.2 & 58.1 & 58.2 & 58.1 & 58.1 & 58.0 & 58.2 & 58.1 & 58.2 & 58.1 & 57.9 \\
    \hline
    \end{tabular}}
    \caption{Q\&A accuracy of Adaptive RMU Zephyr-7B models on WMDP-Biology, WMDP-Cyber, and MMLU. Models were fine-tuned on WMDP-Biology and WMDP-Cyber retain sets.}
    \label{tab6}
\end{table*}

\paragraph{$n$-gram overlap analysis.}  
\begin{figure*}
\centering 
\subfigure[Distribution of Unigram overlap score between WMDP-Biology retain and WMDP-Biology forget sets.]{\includegraphics[width=0.475\textwidth]{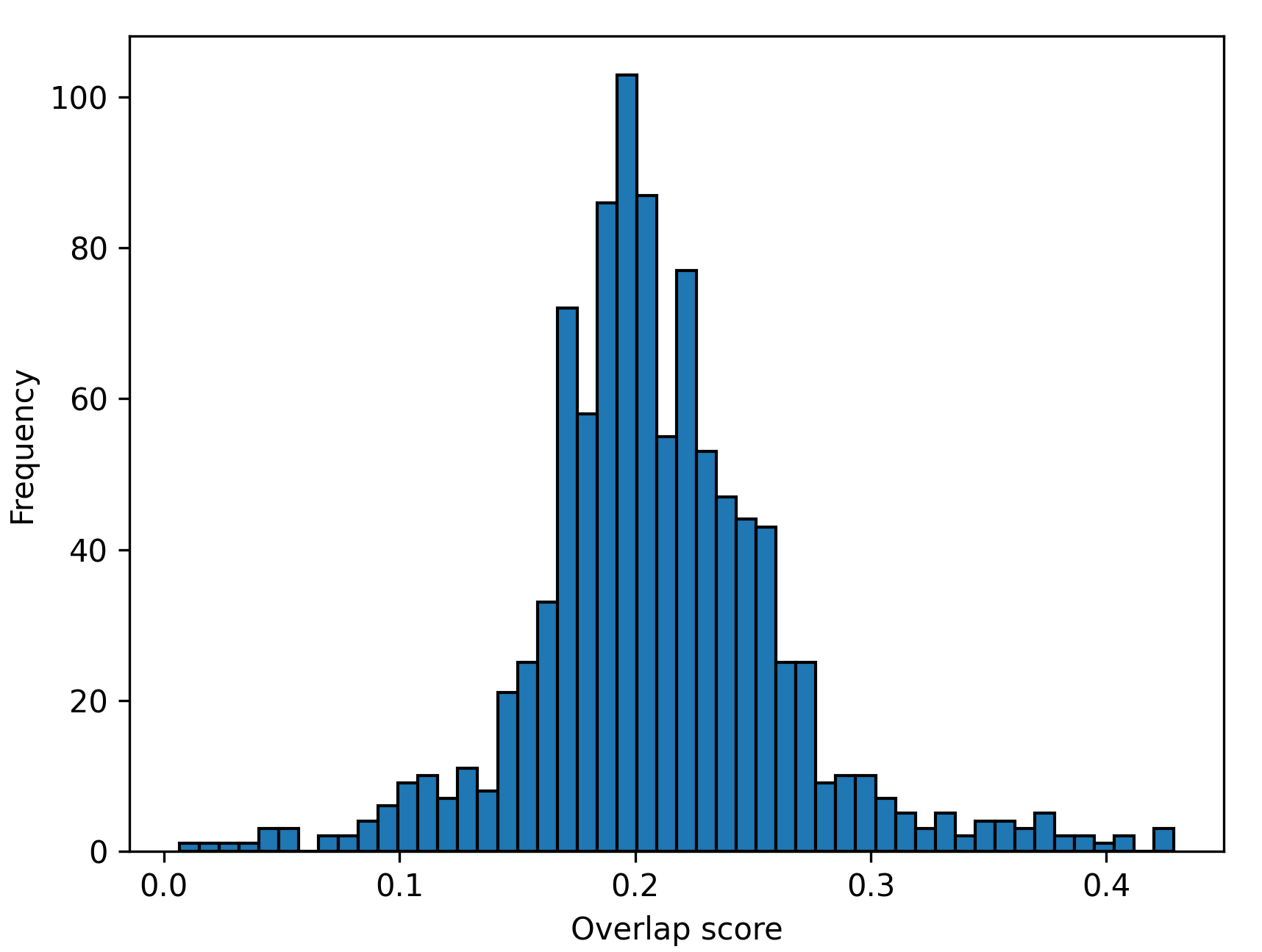}
}\label{fig5a}
\subfigure[Distribution of Bigram overlap score between WMDP-Biology retain and WMDP-Biology forget sets.]{\includegraphics[width=0.475\textwidth]{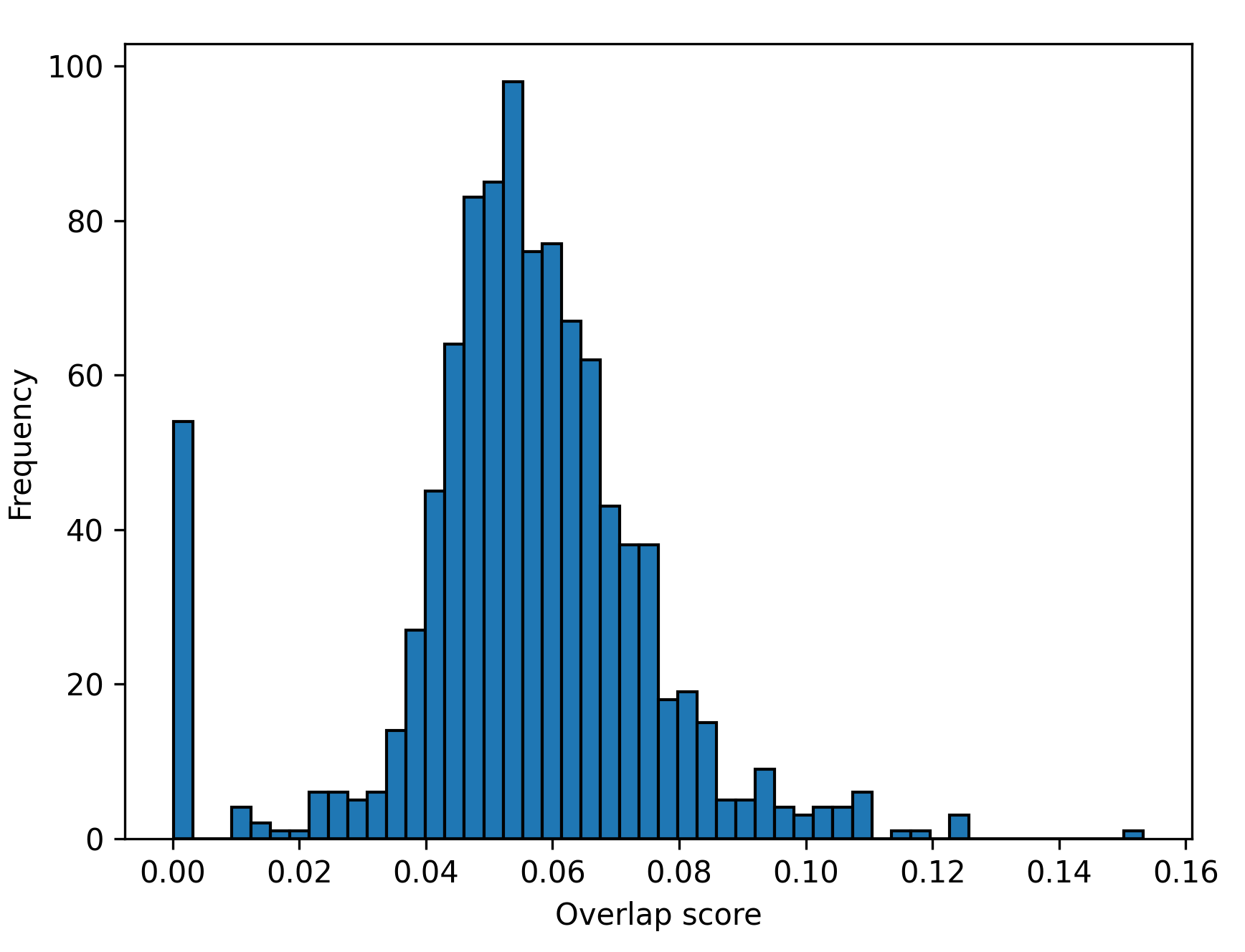}
}\label{fig5b}
\subfigure[Distribution of Unigram overlap score between WMDP-Cyber retain and WMDP-Cyber forget sets.]{\includegraphics[width=0.475\textwidth]{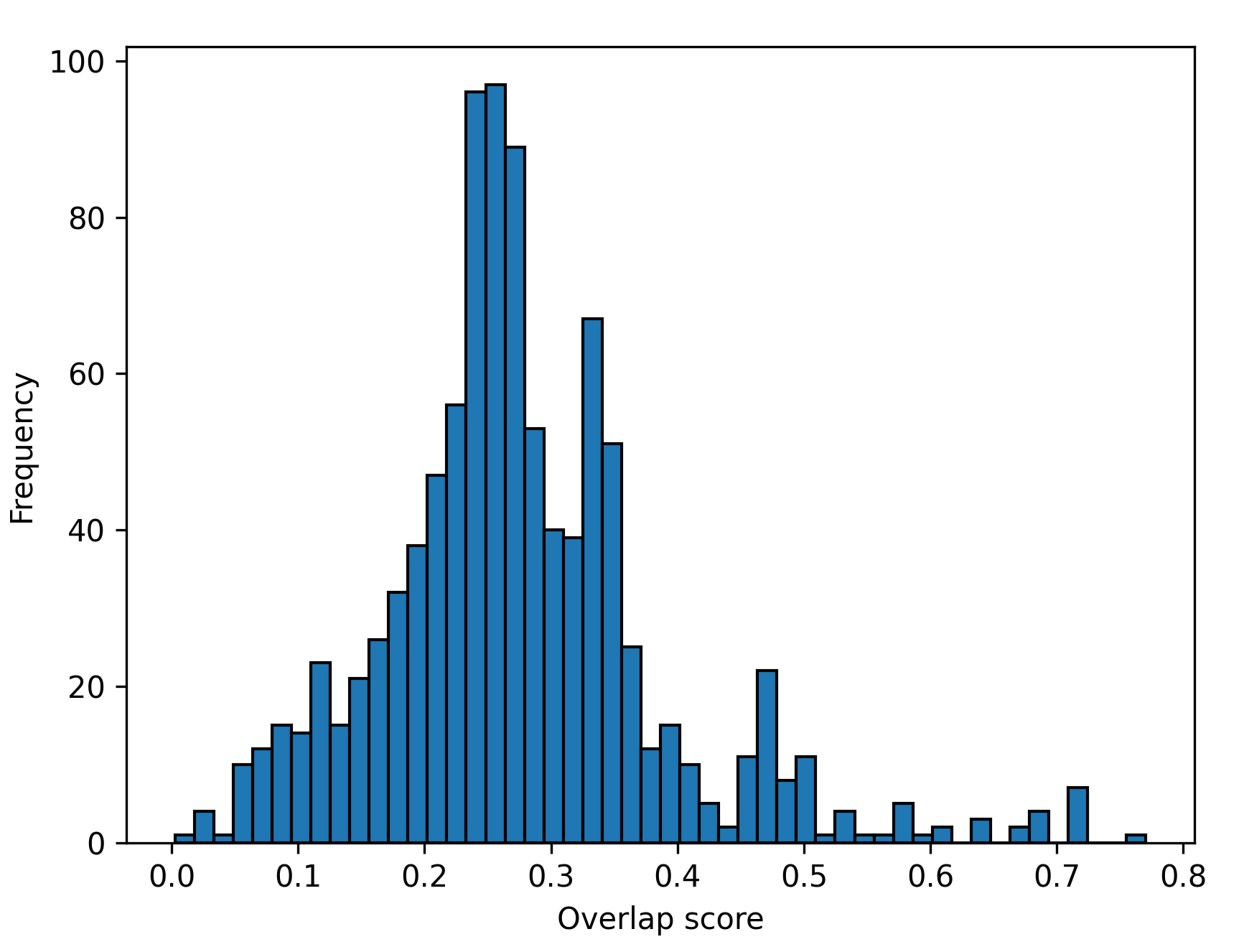}
}\label{fig5c}
\subfigure[Distribution of Bigram overlap score between WMDP-Cyber retain and WMDP-Cyber forget sets.]{\includegraphics[width=0.475\textwidth]{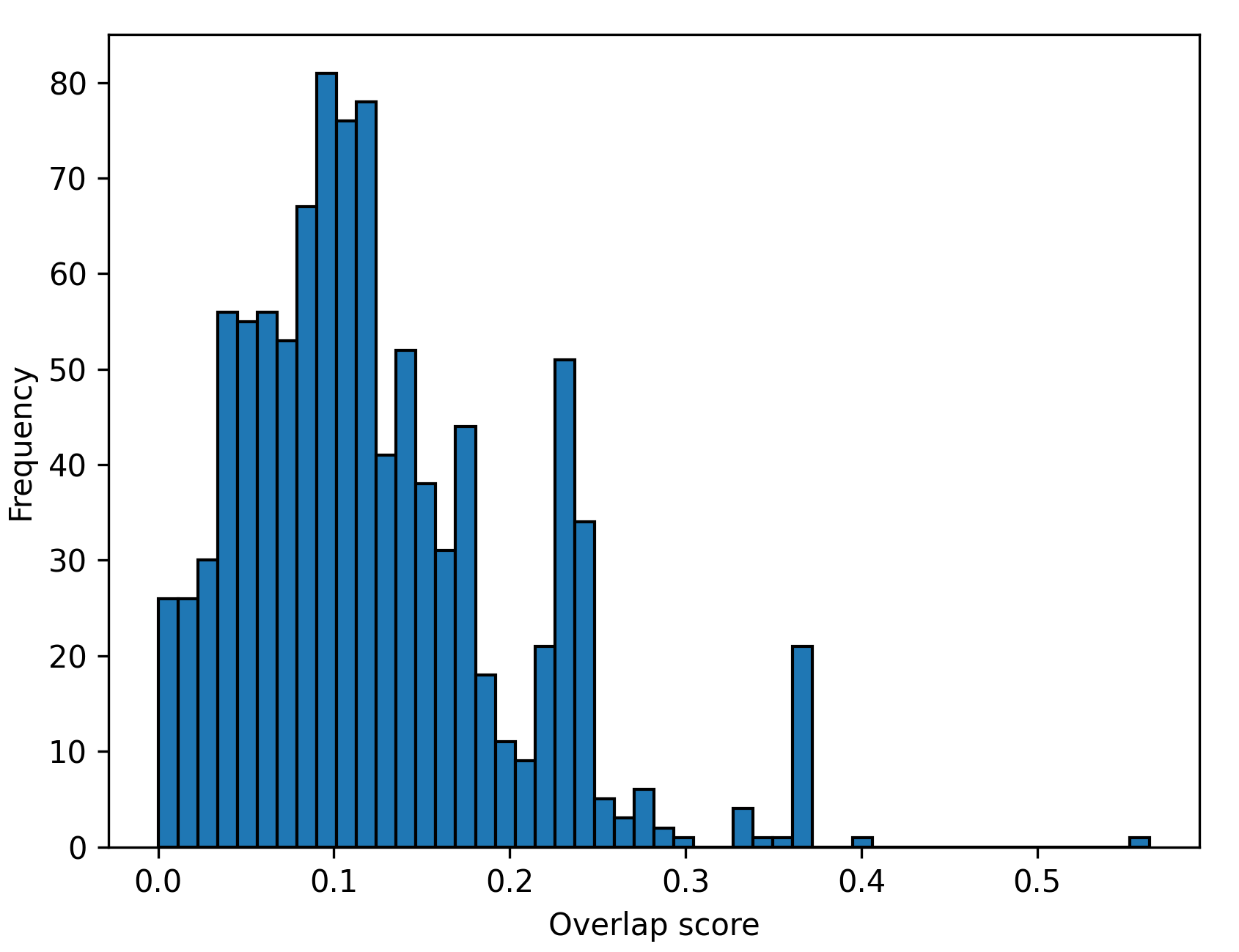}
}\label{fig5d}
\caption{Distributions of Unigram and Bigram overlap scores.}
\label{fig6}
\end{figure*}
Given a retain sample $x_{1:k} \in \mathcal{D}_{\textnormal{retain}}$ consists of $k$ tokens $\{x_1,x_2,...x_k\}$, we denote $x_{i:i+n-1}$ for $i \in [1,...,
k-n+1]$ as the $n$-gram of $x_{1:k}$. The $n$-gram overlap score of $x_{1:k}$ in forget set $\mathcal{D}_{\textnormal{forget}} = \{x_F\}^{|\mathcal{D}_{\textnormal{forget}}|}$ is defined as:
\begin{align}
    \frac{1}{|\mathcal{D}_{\textnormal{forget}}|}\frac{1}{k-n+1}\sum_{x_R}\sum_{i=1}^{k-n+1} \mathbb{I}[x_{i:i+n-1}\in x_F],
\end{align}
where $\mathbb{I}(\cdot)$ is the indicator function and $\mathbb{I}[x_{i:i+n-1}\in x_F] = 1$ if the substring $x_{i:i+n-1}$ is in forget sample $x_F$, otherwise $0$. We randomly sampled $1000$ documents from each dataset and performed Unigram ($n=1$) and Bigram ($n=2$) overlap analysis. The results indicate a high degree of unigram and bigram overlap between the WMDP-forget and WMDP-retain sets. Specifically, the average Unigram and Bigram overlap scores for the WMDP-Biology forget and retain sets were $20.8\%$ and $5.5\%$, respectively. These overlap scores were even higher for the WMDP-Cyber sets, at $27.5\%$ and $12.3\%$, respectively. The distributions of $n$-gram overlap scores are visualized in Fig.~\ref{fig6}. High $n$-gram overlap scores make two distributions WMDP-forget set and WMDP-retain set less distinction, which makes the unlearning more difficult.

\subsection{Limitation and future work}
We discuss the following limitations in our paper: 
\begin{enumerate}
    \item We mainly perform experiments on 7B versions (or equivalent) due to computational constraints. To validate the generalizability  of our approach and findings, we conducted experiments across the Zephyr, Mistral, Llama, and Yi models.
    \item Our analysis in Section~\ref{sec:3.3} on white-box attacks for open weight models. In practice, state-of-the-art  LLMs such as GPT, Gemini, and Claude are trained privately and are accessible through API only. The most common form of attack on LLMs, therefore, is a black-box jailbreak attack. We encourage future works to explore the analysis of the robustness of unlearned models covering black-box jailbreak attacks.
    \item Limiting update the model parameters w.r.t three layer $\{l, l - 1, l - 2\}$ thus risks missing interesting generalization behaviors.

\end{enumerate}


\end{document}